%% file: main.tex
\documentclass[acmtog]{acmart}

\usepackage{color}
\definecolor{bittersweet}{rgb}{1.0, 0.44, 0.37}

\usepackage{booktabs} 
\usepackage{wrapfig}
\usepackage{overpic}
\usepackage{multirow}

\newcommand{\under}[1]{\underline{\textbf{#1}}}

\usepackage[ruled]{algorithm2e} 

\SetAlFnt{\small}
\SetAlCapFnt{\small}
\SetAlCapNameFnt{\small}
\SetAlCapHSkip{0pt}

\begin{document}
\title{NeurCADRecon: Neural Representation for Reconstructing CAD Surfaces by Enforcing Zero Gaussian Curvature}

\author{Qiujie Dong}
    \orcid{0000-0001-6271-2546}
    \affiliation{%
    \institution{Shandong University}
    \city{Qingdao}
    \state{Shandong}
    \country{China}}
    \email{qiujie.jay.dong@gmail.com}
    
\author{Rui Xu}
    \orcid{0000-0001-8273-1808}
    \affiliation{%
    \institution{Shandong University}
    \city{Qingdao}
    \state{Shandong}
    \country{China}}
    \email{xrvitd@163.com}
    
\author{Pengfei Wang}
    \orcid{0000-0002-0938-267X}
    \affiliation{%
    \institution{The University of Hong Kong}
    \state{Hong Kong}
    \country{China}}
    \email{pf.wang.graphics@qq.com}
    
\author{Shuangmin Chen}
    \orcid{0000-0002-0835-3316}
    \affiliation{%
    \institution{Qingdao University of Science and Technology}
    \city{Qingdao}
    \state{Shandong}
    \country{China}}
    \email{csmqq@163.com}
    
\author{Shiqing Xin}
\authornote{Corresponding author: Shiqing Xin.}         \orcid{0000-0001-8452-8723}
    \affiliation{%
    \institution{Shandong University}
    \city{Qingdao}
    \state{Shandong}
    \country{China}}
    \email{xinshiqing@sdu.edu.cn}
    
\author{Xiaohong Jia}
    \orcid{0000-0001-6206-3216}
    \affiliation{%
    \institution{AMSS, Chinese Academy Of Sciences}
    \state{Beijing}
    \country{China}}
    \email{xhjia@amss.ac.cn}
    
\author{Wenping Wang}
    \orcid{0000-0002-2284-3952}
    \affiliation{%
    \institution{Texas A\&M University}
    \state{Texas}
    \country{USA}}
    \email{wenping@tamu.edu}
    
\author{Changhe Tu}
    \orcid{0000-0002-1231-3392}
    \affiliation{%
    \institution{Shandong University}
    \city{Qingdao}
    \state{Shandong}
    \country{China}}
    \email{chtu@sdu.edu.cn}

\input{sections/00-abstract}

\input{sections/01-intro}

\input{sections/02-related}
\input{sections/03-method}

\input{sections/04-exp}

\input{sections/05-ablation}
\input{sections/06-conclusion}

\section*{Acknowledgments}
The authors would like to thank the anonymous reviewers for their valuable comments and suggestions. This work is supported by National Key R\&D Program of China (2021YFB1715900), National Natural Science Foundation of China (62272277, U23A20312, 62072284) and NSF of Shandong Province (ZR2020MF036).

\bibliographystyle{ACM-Reference-Format}
\bibliography{references}

\end{document}

%% file: sections/00-abstract.tex
\begin{abstract}
\label{sec:abstract}
Despite recent advances in reconstructing an organic model with the neural signed distance function (SDF), the high-fidelity reconstruction of a CAD model directly from low-quality unoriented point clouds remains a significant challenge. In this paper, we address this challenge based on the prior observation that the surface of a CAD model is generally composed of piecewise surface patches, each approximately developable even around the feature line. Our approach, named~{\em NeurCADRecon}, is self-supervised, and its loss includes a developability term to encourage the Gaussian curvature toward 0 while ensuring fidelity to the input points (see the teaser figure). Noticing that the Gaussian curvature is non-zero at tip points, we introduce a double-trough curve to tolerate the existence of these tip points. Furthermore, we develop a dynamic sampling strategy to deal with situations where the given points are incomplete or too sparse. Since our resulting neural SDFs can clearly manifest sharp feature points/lines, one can easily extract the feature-aligned triangle mesh from the SDF and then decompose it into smooth surface patches, greatly reducing the difficulty of recovering the parametric CAD design. A comprehensive comparison with existing state-of-the-art methods shows the significant advantage of our approach in reconstructing faithful CAD shapes.
\end{abstract}

\ccsdesc[500]{Computing methodologies~Mesh models}

\ccsdesc[500]{Computing methodologies~Point-based models}

\ccsdesc[500]{Computing methodologies~Mesh geometry models}

%
%

\keywords{
CAD model,
unoriented point cloud,
surface reconstruction,
signed distance function,
Gaussian curvature
}

\begin{teaserfigure}
    \centering
    \includegraphics[width=\textwidth]{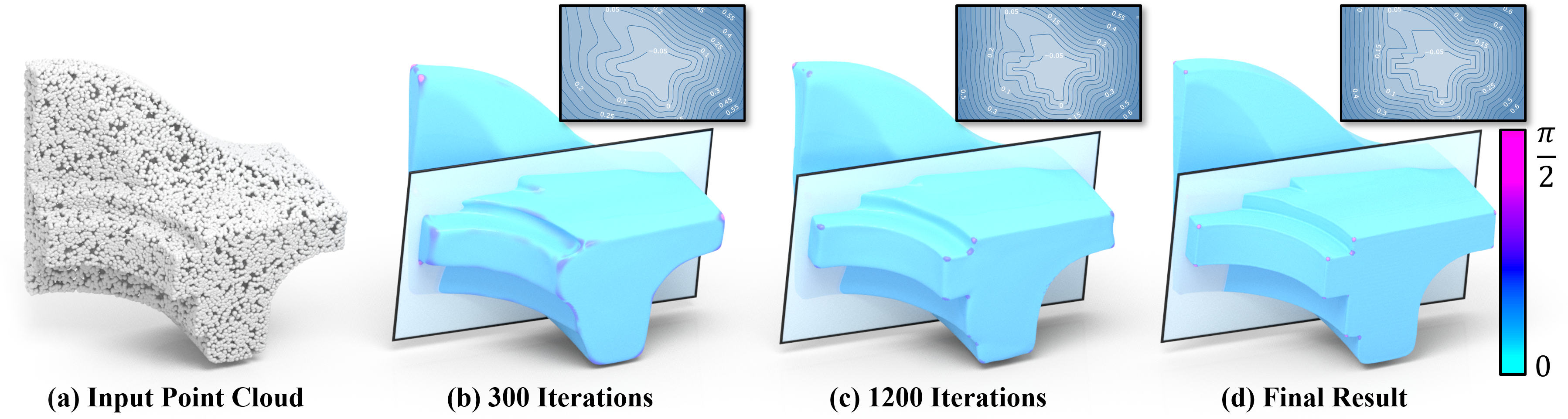}
    \vspace{-6mm}
    \caption{In this paper, we propose a self-supervised deep neural network for reconstructing a CAD model from an unoriented point cloud. Considering that the surface of a CAD model is piecewise smooth, with each surface patch being approximately developable, we encourage the Gaussian curvature toward 0 while ensuring fidelity to the input points in the process of optimizing the signed distance function (see the close-up windows for the SDF visualization in a sectional plane). Tests on four public datasets show that our approach consistently outperforms the state-of-the-art methods, thanks to the utilization of the prior. From left to right: a point cloud with 10K points and the intermediate reconstruction results with color-coded visualization of absolute Gaussian curvatures (300 iterations, 1.2K iterations, and 9K iterations, respectively).}
    \label{fig:teaser}
\end{teaserfigure}

\maketitle

%% file: sections/01-intro.tex
\section{Introduction}
\label{sec:intro}
Reverse engineering of CAD products is a fundamental yet challenging task with diverse applications in both academic and industrial communities~\cite{CAPRI_Net}. While significant progress has been made in reconstructing an organic model directly from a low-quality unoriented point cloud~\cite{IGR, SIREN, IDF, Rui2022RFEPS}, the high-fidelity reconstruction of a CAD model remains a significant challenge. This challenge arises from the fact that the surface of a CAD model may contain sharp feature points/lines, posing a considerable obstacle when encoding the shape using a neural signed distance function~(SDF).
\begin{wrapfigure}{r}{4cm}
\vspace{0mm}
  \hspace*{-6mm}
  \centerline{
  \includegraphics[width=45mm]{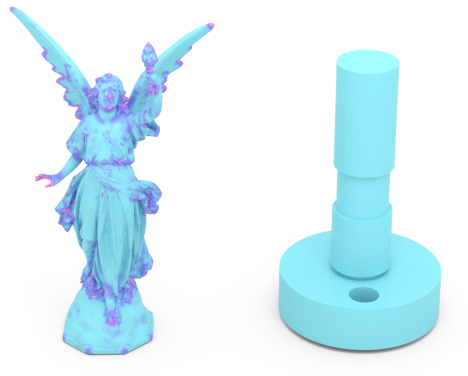}
  }
  \vspace*{-6mm}
\end{wrapfigure}

The prominent characteristic of CAD models lies in that they are commonly piecewise, with each surface patch being smooth and approximately developable, which holds true even for a point on the feature line. Given that developability can be characterized by zero Gaussian curvature, it is reasonable to encourage the Gaussian curvature toward 0 in surface reconstruction, as illustrated in the inset figure depicting Gaussian curvature on the free model and CAD model. Based on this observation, we propose a self-supervised neural network to learn the signed distance function (SDF) of the underlying surface. In implementation, we inherit the Dirichlet condition~\cite{phase} and the Eikonal condition~\cite{IGR} to ensure fidelity while adding a loss term to minimize the overall absolute Gaussian curvature. With increasing iterations, the SDF can not only respect the input point cloud but also manifest the sharp feature points. See the teaser figure for an illustration.

\begin{figure}[t]
    \centering
    \begin{overpic}[width=0.95\linewidth]{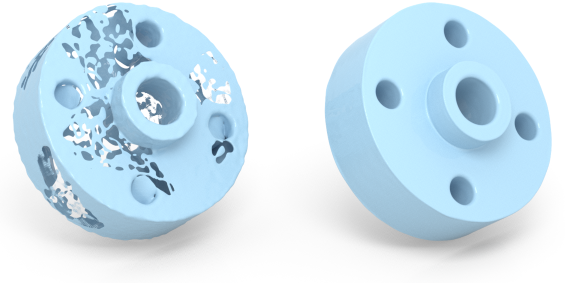}
\put( 6,1.5){\textbf{Rank of Hessian = 1 }}
\put(60,1.5){\textbf{Gaussian curvature = 0}}
    \end{overpic}
    \vspace{-4mm}
    \caption{
   Despite being mathematically correct, enforcing the rank of the Hessian matrix to be 1 may lead to numerical instability. In this paper, we propose to reconstruct CAD shapes by minimizing the overall absolute Gaussian curvature.
    }
    \label{fig:min_rank_recon}
    \vspace{-6mm}
\end{figure}

Suppose that $p$ is a surface point with a normal vector~$\boldsymbol{n}_p$, where the two principal curvature directions are $\boldsymbol{\alpha}_p$ and $\boldsymbol{\beta}_p$ with the corresponding principal curvatures being $\kappa_1$ and $\kappa_2$, respectively. According to the well-established relationship between curvatures and the Hessian matrix of the SDF, $\boldsymbol{n}_p$, $\boldsymbol{\alpha}_p$, and $\boldsymbol{\beta}_p$ define the three eigenvectors of the Hessian matrix $H$ with at least two zero eigenvalues, one of which corresponds to $\boldsymbol{n}_p$ and the other corresponds to $\boldsymbol{\alpha}_p$ or $\boldsymbol{\beta}_p$. Because of this, a traditional formulation of zero Gaussian curvature is achieved by enforcing the rank of~$H$ to be at most~1.0~\cite{Develop2020}. Despite its correctness in mathematics, it fails to consistently report a reasonable solution due to numerical instability; See Fig.~\ref{fig:min_rank_recon} for the difference. We shall discuss this problem in Section~\ref{subsec:Constraint}.

In addition, the implementation process requires addressing at least three challenges. Firstly, as depicted in the teaser figure, the Gaussian curvature significantly deviates from 0 at tip points. Typical corner points display a Gaussian curvature of approximately $\pi/2$. Imposing a universal constraint of 0 may result in an undesirable bulge around the tip points. To address this, we introduce a double-trough curve to accommodate the presence of tip points.
Secondly, considering that some CAD surfaces may have non-developable surface patches, we employ the technique of learning rate annealing to gradually reduce the influence of developability over time.
Lastly, the input points may exhibit various imperfections, such as missing parts or high sparsity. If the loss is measured solely around the input point cloud, it is unlikely to enforce the constraints over the entire surface. Therefore, we implement a dynamic sampling strategy, enabling adaptive loss measurement that varies with changes in the underlying surface.

\begin{figure*}[!tp]
    \centering
    \includegraphics[width=0.98\linewidth]{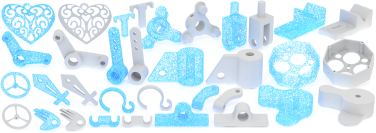}
    \vspace{-4mm}
    \caption{
    A gallery of reconstruction results by our NeurCADRecon. The central idea is to encourage the Gaussian curvature toward 0 while ensuring fidelity to the input points.
    }
    \label{fig:gallery}
\end{figure*}

Our contributions are three-fold:
\begin{enumerate}
   \item We propose to learn the neural SDF, in a self-supervised fashion, from a low-quality unoriented point cloud that potentially represents a CAD model, inspired by the fact that CAD model surfaces are typically piecewise smooth and approximately developable.
   
    \item We define the developability loss term for encouraging the target Gaussian curvature to be either 0 or around $\pi/2$ (tolerating the existence of tip points), which is achieved by introducing a double-trough curve.
    
    \item We leverage dynamic sampling to handle data imperfections, 
    allowing for adaptive loss measurement varying with changes in the underlying surface. Extensive tests on four open datasets showcase our significant advantage in reconstructing CAD-type shapes; See Fig.~\ref{fig:gallery} for a gallery of reconstruction results.
\end{enumerate}

%% file: sections/02-related.tex
\section{Related Work}
\label{sec:related}
This paper focuses on reconstructing CAD-type point clouds within the surface reconstruction category. The section commences with a review of traditional reconstruction approaches, proceeds to explore deep learning-based methods, and concludes by addressing surface reconstruction under the assumption that the underlying surface represents a CAD model.

\subsection{Traditional Approaches}
Traditional implicit reconstruction approaches, extensively studied in computer graphics~\cite{kolluri2008imls, oztireli2009RIMLS, shen2004interpolating, schroers2014HessianIMLS, MPU, SSD, GCNO}, have played a significant role in this field. \citet{carr2001reconstruction}~pioneered the use of radial basis functions (RBF) to fit the zero-isosurface of the Signed Distance Function (SDF), enabling the reconstruction of smooth surfaces. This approach saw later refinements by~\citet{VIPSS} and~\citet{li2016sparse}.
Furthermore, Kazhdan et al.~\shortcite{Kazhdan2006PoissonSR, SPSR, Kazhdan2020PoissonSR} introduced Poisson surface reconstruction (PSR) and its subsequent variants, formulating the problem of computing the underlying occupancy field as a Poisson equation. However, PSR-based methods are limited to oriented points. To address this limitation, \citet{iPSR} presented an improved PSR (iPSR) method, iteratively taking the normals obtained from the current reconstructed surface as input for the next iteration, completely eliminating the dependence on point normals.
Parametric Gauss Reconstruction (PGR)~\cite{PGR} focuses more on the global consistency of normal orientations, building on the Gauss formula in potential theory to infer the occupancy field. However, none of the mentioned approaches are specifically designed for CAD-type point clouds.

\subsection{Learning-Based Approaches}
As with other learning-based tasks, implicit surface reconstruction methods encompass both supervised and self-supervised approaches.

\paragraph{Supervision Based}
By leveraging prior information provided by ground-truth shape assemblies, supervision-based implicit surface reconstruction methods~\cite{ONet, DeepLS, jiang2020local, SA-ConvONet, DeepSDF, Points2surf, tang2021octfield, wang2022dual, lin2023patchgrid} aim to learn the implicit representation (e.g., SDF or occupancy field) of the given point cloud. \citet{SAP} introduced a differentiable point-to-mesh layer using a differentiable formulation of Poisson Surface Reconstruction (PSR), enabling a GPU-accelerated fast solution for the indicator function. This concept was further extended by \citet{GalerkinNG} to incorporate learnable basis functions. Additionally, ConvONet~\cite{Peng20ConvONet} employed a convolutional architecture to map 3D points to a feature grid, initially designed for predicting occupancy fields. Furthermore, \citet{POCO} enhanced the quality and performance of ConvONet~\cite{Peng20ConvONet} by replacing convolutions with a transformer architecture.

\paragraph{Self-supervision Based}
Self-supervised implicit surface reconstruction methods~\cite{CAP-UDF, PCP} are often referred to as fitting-based methods. These approaches directly fit the implicit representation from raw point clouds without relying on any ground truth shape assemblies, thus enhancing their generalization capability.
SAL~\cite{SAL, SALD} introduces Sign Agnostic Learning (SAL), characterized by a family of loss functions that can be applied directly to unsigned geometric data. This approach produces signed implicit representations of surfaces.
IGR~\cite{IGR} utilizes the Eikonal term to achieve implicit geometric regularization, offering an effective method for reconstructing surfaces.
SIREN~\cite{SIREN} achieves impressive reconstruction results by utilizing periodic activation functions. These functions preserve high frequencies while maintaining low-frequency implicit representations through the bias of ReLU-MLPs.
Inspired by SIREN, \citet{yifan2020isopoints} proposed a hybrid neural surface representation. This method focuses on imposing geometry-aware sampling and regularization, using iso-points as an explicit representation for a neural implicit function.
Neural-Pull~\cite{NeuralPull} advocates for predicting both the signed distance and gradients simultaneously. This allows a query point to be pulled to the nearest point on the underlying surface, enhancing accuracy.
DiGS~\cite{DiGS} applies the Laplacian energy as a soft constraint of the Signed Distance Function (SDF). This method is effective for reconstructing surfaces from unoriented point clouds.
Lastly, Neural-Singular-Hessian~\cite{HessianZX} enforces the Hessian of the neural implicit function to have a zero determinant for points near the surface. This approach is particularly useful for recovering details from unoriented point clouds.

\subsection{Surface Reconstruction of CAD Models}
In addressing the challenge of reconstructing CAD-type models, there has been significant prior research, notably by~\citet{CSGNet} and~\citet{UCSG}, focusing on primitive fitting.
\citet{SPFN} employed supervised learning to initially detect primitive types and fit various primitive patches, including planes, cylinders, cones, and spheres.
\citet{ParseNet} extended SPFN's approach by incorporating B-spline patches and integrating differentiable metric-learning segmentation, representing a step forward in the field.
\citet{Point2Cyl} segmented raw 3D point clouds into sets of extrusion cylinders, using boolean combination operations to merge these primitives, presenting a novel approach to the problem.
\citet{CAPRI_Net} introduced an unsupervised approach that similarly involved primitive prediction and boolean combination operations. However, both of the approaches~\cite{Point2Cyl, CAPRI_Net} favor voxelized inputs over point clouds. This necessitates the voxelization of oriented point clouds before they are input into the neural network, or before being used to learn a latent code.

It is evident that directly using specific primitives to fit a complex CAD model from a low-quality point cloud may not be practical. Therefore, we propose an alternative: fitting a high-fidelity neural Signed Distance Function (SDF) that can clearly manifest distinct feature points and lines. With a geometry that is sufficiently faithful, it becomes straightforward to discretize it into a set of surface patches and recover the parametric design. For an illustrative example, see Section~\ref{sec:compare_dataset}.

%% file: sections/03-method.tex
\section{Our approach}
\label{sec:method}
As with other existing self-supervision based methods, our approach, named \textit{NeurCADRecon}, takes an unoriented point cloud as input and focuses on learning the neural Signed Distance Function (SDF) to accurately represent the underlying surface. The design of NeurCADRecon is centered around two key aspects. Firstly, it inherits the Dirichlet condition~\cite{phase} and the Eikonal condition~\cite{IGR} to ensure high fidelity to the input points. Secondly, NeurCADRecon emphasizes encouraging the Gaussian curvature towards zero, which is crucial for highlighting the sharp features characteristic of CAD models.

\subsection{Preliminaries}

We assume that the input is a point cloud, $\mathcal{P}$, which does not include normals. The goal is to predict a neural Signed Distance Function (SDF), denoted as $f(\boldsymbol{x};\Theta):\mathbb{R}^3\to\mathbb{R}$, such that its zero level-set accurately represents the underlying surface, $\mathcal{S}$. Here, $\Theta$ symbolizes the computational parameters of a network that necessitate frequent optimization and updates.

Despite their differences, existing approaches primarily concentrate on developing a geometrically meaningful loss function to aid in discovering the desired solution. Given that $\boldsymbol{p} \in \mathcal{P}$ is considered to be on the surface $\mathcal{S}$, it is often labeled as a manifold point. Moreover, a distinct set of sample points, $\mathcal{Q}$, usually uniformly sampled from its bounding box (normalized to the range~$[-0.5, 0.5]^3$ by default), is employed to help regularize the characteristics of the underlying SDF. In the following, we provide a brief overview of the Dirichlet condition and the Eikonal condition. These conditions have been extensively discussed in the research community and have proven to be valuable in works such as~\citet{IGR}, \citet{SIREN}, \citet{SAL}, \citet{NeuralPull}, and~\citet{DiGS}.

\paragraph{Eikonal Condition}
As a valid SDF, 
$f(\boldsymbol{x};\Theta)$ must possess a unit gradient for each point $\boldsymbol{p} \in \mathcal{P}$,
i.e., that is~$\| {\nabla_{\boldsymbol{p}}} f\|=1$.
Based on this requirement, the Eikonal condition can be written as a loss term:
\begin{equation}
    \mathcal{L}_{E} = \frac{1}{|\mathcal{P}|+|\mathcal{Q}|}\int_{\mathcal{P}\cup\mathcal{Q}}\big\vert  1 - \| \nabla f(\boldsymbol{x};\Theta)\| \big\vert \text{d}\boldsymbol{x},
\end{equation}
where $|\cdot|$ is the operator for taking the number of elements.

\paragraph{Dirichlet Condition}
For any point $\boldsymbol{p}$ within $\mathcal{P}$, it is crucial that they align closely with the reconstructed surface, ideally satisfying $f(\boldsymbol{p};\Theta) = 0$ as much as possible. This alignment ensures that the points from the input cloud are accurately represented on the surface. Conversely, for any point $\boldsymbol{q}$ in $\mathcal{Q}$, we regard it as external to the underlying surface, hence we encourage $f(\boldsymbol{q};\Theta) \neq 0$. This helps regularize the shape by delineating the surface from the surrounding space. 
They can be respectively written as loss terms:
\begin{equation}
    \mathcal{L}_\text{DM} = \frac{1}{|\mathcal{P}|}\int_{\mathcal{P}}{\big\vert f(\boldsymbol{p};\Theta) \big\vert}\text{d}\boldsymbol{p},
\end{equation}
and
\begin{equation}
    \mathcal{L}_\text{DNM} = \frac{1}{|\mathcal{Q}|}\int_{\mathcal{Q}}{exp(-\alpha\big\vert f(\boldsymbol{q};\Theta) \big\vert)}\text{d}\boldsymbol{q}.
\end{equation}

Given that the input point cloud lacks normals, there is no need to impose constraints on normal vectors. In light of this, the loss function comprises three terms as below:
\begin{equation}
\mathcal{L}_\text{basic} = \lambda_\text{E}  \mathcal{L}_\text{E} + \lambda_\text{DM} \mathcal{L}_\text{DM} + \lambda_\text{DNM} \mathcal{L}_\text{DNM}.
\end{equation}
However, $\mathcal{L}_\text{basic}$ fails to adequately capture the sharp feature points and lines that are typically associated with CAD geometries. This limitation underscores the need for integrating additional components into the loss function, aimed at enhancing the reconstruction fidelity of CAD models. 

\subsection{Gaussian Curvature Constraint Term}
\label{subsec:Constraint}
We observe that the surface of a CAD model is commonly piecewise smooth and approximately developable. Given that developability implies zero Gaussian curvature, we introduce an additional loss term to specify this requirement. The proposed loss term is formulated as follows:
\begin{equation}
    \mathcal{L}_\text{Gauss}^{(1)} = \frac{1}{|\mathcal{P}|}\int_{\mathcal{P}}{\big\vert k_\text{Gauss}(\boldsymbol{x}) \big\vert}\text{d}\boldsymbol{x}.
\end{equation}

\begin{wrapfigure}{r}{2.5cm}
\vspace{-3mm}
  \hspace*{-4mm}
  \centerline{
  \includegraphics[width=30mm]{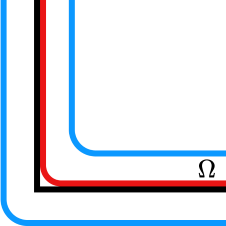}
  }
  \vspace*{-4mm}
\end{wrapfigure}
Recall that $f(\boldsymbol{x};\Theta)$ aims to approximate the actual SDF and demonstrates differentiability throughout the entire space. Suppose we aim to reconstruct a target surface resembling the cube model depicted in the inset figure. The real SDF might not be differentiable within a sufficiently narrow space surrounding the original surface due to sharp feature lines. However, by slightly smoothing these sharp feature lines, we can establish a narrow differential space~$\Omega$. This adjustment allows us to leverage the differential properties of $f(\boldsymbol{x};\Theta)$ to estimate various geometric quantities effectively.

For example, for a point~$\boldsymbol{x} \in \Omega$, the Gaussian curvature of $\boldsymbol{x}$
can be estimated through the hessian~$H_f(\boldsymbol{x})$~\cite{Spivak1979GaussianCurvature, knoblauch1913GaussianCurvature, Ron2005GaussianCurvature}:
\begin{equation}
\label{eq:det_gaussian_curvature}
    k_\text{Gauss}(\boldsymbol{x}) = - \frac
            {\left | \begin{matrix}
                H_f(\boldsymbol{x}) & \nabla f^T(\boldsymbol{x};\Theta) \\
                \nabla f(\boldsymbol{x};\Theta) & 0  \\
                \end{matrix} \right | }
            {\| \nabla f(\boldsymbol{x};\Theta) \|^4}.
\end{equation}

\paragraph{Comparison with Rank Constraint}
Suppose $p$ is a surface point with a normal vector $\boldsymbol{n}_p$, and the two principal curvature directions at $p$ are $\boldsymbol{\alpha}_p$ and $\boldsymbol{\beta}_p$, with the corresponding principal curvatures being $\kappa_1$ and $\kappa_2$, respectively. The eigenvectors of the Hessian matrix $H$ are represented by $\boldsymbol{n}_p$, $\boldsymbol{\alpha}_p$, and $\boldsymbol{\beta}_p$. Firstly, the eigenvalue corresponding to $\boldsymbol{n}_p$ is 0~\cite{HessianZX}. Secondly, one of $\kappa_1$ and $\kappa_2$ must be 0 due to zero Gaussian curvature, i.e., $\kappa_1\times\kappa_2 = 0$. This indicates that the rank of $H_f$ is at most 1, as discussed in~\cite{Develop2020}. Such a rank constraint has been utilized in achieving developability and fitting algebraic spline surfaces.

Despite the mathematical equivalence, the constraint of the rank being at most~1 faces numerical challenges. Firstly, a slight change in $H_f$ can cause a jump in the rank, hinting at potential numerical instability. Secondly, when the target shape resembles a cube-like model, the neural SDF $f$ cannot exactly match the actual SDF, as the latter is not differentiable even in a very narrow space around the surface.
The neural SDF $f$, despite being slightly from the actual SDF, 
has a narrow space~$\Omega$ surrounding the surface.
Although it is convenient to estimate Gaussian curvature through the neural SDF~$f$ (slightly from the actual SDF) and its Hessian matrix~$H_f$, 
the rank of~$H_f$ may not reflect the real situation. 
Therefore, in this paper, we propose minimizing the overall absolute Gaussian curvature to achieve the goal of developability.

\paragraph{Sampling Strategy for Representing~$\Omega$}
Clearly, it is unfeasible to consider the contribution of every point in $\Omega$ directly. Following the practices of works like~\citet{IGR}, \citet{NeuralPull}, and \citet{CAP-UDF}, we utilize a sampling strategy that involves Gaussian distributions centered on each input point $\boldsymbol{p} \in \mathcal{P}$ to select points from $\Omega$. To elaborate, for a given point $\boldsymbol{p} \in \mathcal{P}$, the Gaussian distribution is centered at $\boldsymbol{p}$, with its mean and standard deviation determined by the distance to its $k$-th nearest neighbor, where $k$ is set to 50 by default. This method employs a one-point sampling technique for each distribution to maintain a consistent number of points sampled from $\Omega$~(15K by default) aligning with the batch size used in training.
By abusing notations, we reformulate the Gaussian curvature term as follows:
\begin{equation}
    \mathcal{L}_\text{Gauss}^{(2)} = \frac{1}{|{\Omega}|}\int_{{\Omega}}{\big\vert k_\text{Gauss}(\boldsymbol{x}) \big\vert}\text{d}\boldsymbol{x}.
\end{equation}

\subsection{Double-trough Function}
\begin{figure}[tp]
    \vspace{3mm}
    \centering
    \begin{overpic}[width=0.95\linewidth]{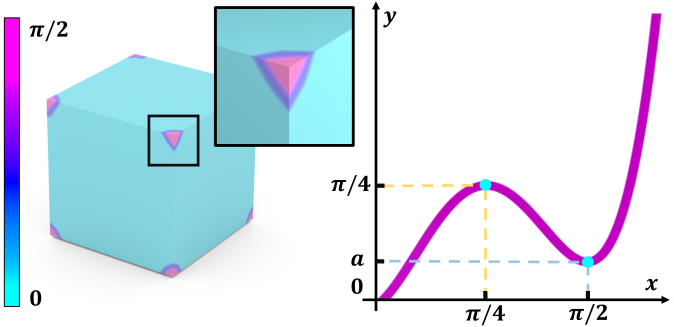}
    \put(23, -3){\textbf{(a)}}
    \put(70, -3){\textbf{(b)}}
    \end{overpic}
    \caption{
(a) The Gaussian curvature field on the cube model shows significant deviations from 0 at the tip points. 
(b) To accommodate the non-zero Gaussian curvature at the tip points, 
we advocate the use of a double-trough function. This function is designed to permit the desired Gaussian curvature to assume values of either 0 or approximately $\pi/2$.
    }
    \label{fig:corner_cube_myFunction}
    \vspace{-5mm}
\end{figure}

As illustrated in Fig.~\ref{fig:corner_cube_myFunction}(a), it is clear that the Gaussian curvature at tip points is non-zero, a characteristic frequently observed in many CAD models. Moreover, the Gaussian curvature at a tip point significantly deviates from 0, typically approaching a value around~$\pi/2$. 
Consequently, we need to tolerate the existence of non-zero Gaussian curvature at the tip points while leaning towards encouraging a Gaussian curvature closer to 0.
For such a purpose, we invent a double-trough function, as is shown in Fig.~\ref{fig:corner_cube_myFunction}(b). 

The key idea involves mapping the Gaussian curvature of around $\pi/2$ to a value close to 0. To achieve this, we define the double-trough curve $\text{DT}(t)$ based on the following requirements:
\begin{equation}
\left\{
\begin{aligned}
    &\text{DT}(0) = 0\\
     &\text{DT}(\pi/4) = \pi/4\\
    &\text{DT}'(\pi/4) = 0\\
    &\text{DT}(\pi/2) = a\\
    &\text{DT}'(\pi/2) = 0.
\end{aligned}
\right.
\end{equation}
We explain the equations as follows.
$\text{DT}'(\pi/4) = 0$ is used to form a peak at $\pi/4$
while $\text{DT}'(\pi/2) = 0$ is used to form a valley at $\pi/2$, with a height of~$a$.
We set $a$ to $1/4$ by default, indicating that while the presence of non-zero Gaussian curvature is allowed, it is more inclined to favor zero Gaussian curvature. 
It is easy to find a quartic function to satisfy the four equations at the same time:
\begin{equation}
  \text{DT}(t)=\frac{64\pi - 80}{\pi^4}t^4 - \frac{64\pi - 88}{\pi^3}t^3 + \frac{16\pi-29}{\pi^2}t^2+\frac{3}{\pi}t.
\end{equation}

To this end, we can reformulate the Gaussian curvature term as follows:
\begin{equation}
    \mathcal{L}_\text{Gauss} = \frac{1}{|{\Omega}|}\int_{{\Omega}}{\text{DT}(\big\vert k_\text{Gauss}(\boldsymbol{x}) \big\vert)}\text{d}\boldsymbol{x}.
\end{equation}

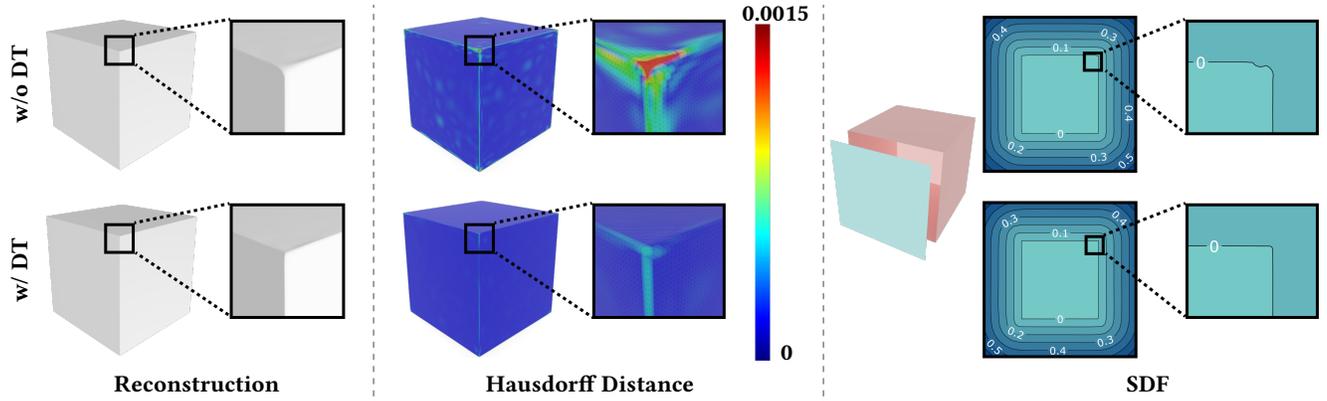
\begin{figure*}[!tp]
    \vspace{2mm}
    \centering
        \begin{overpic}[width=0.95\linewidth]{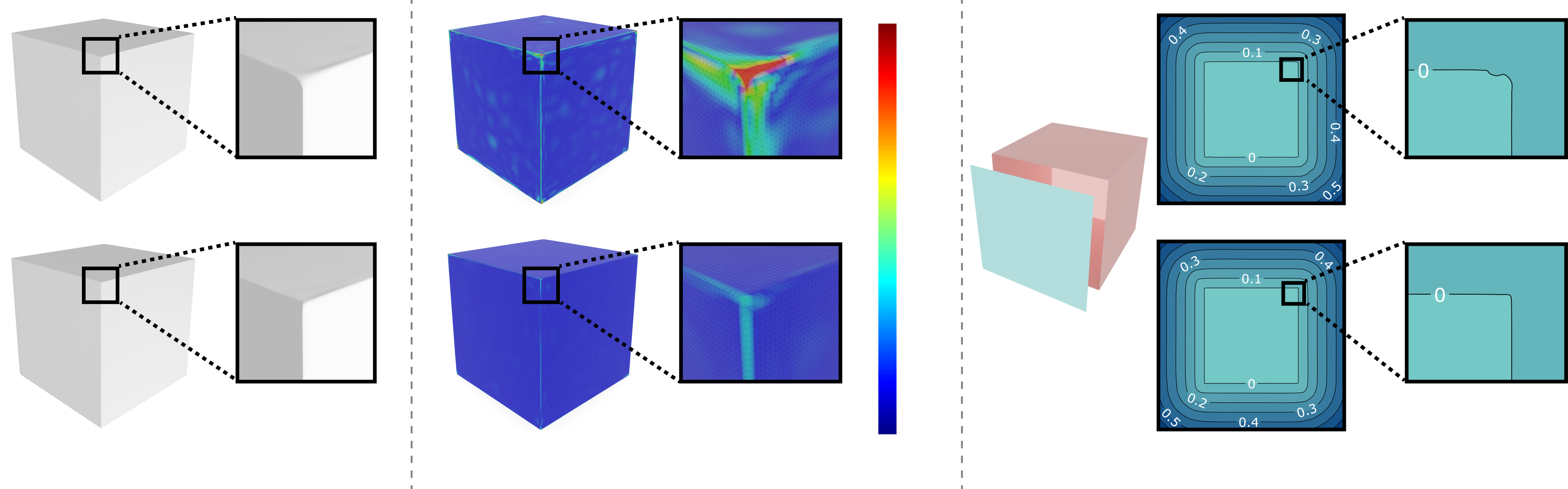}
    \put( -2, 8){\rotatebox{90}{\textbf{w/ DT}}}
    \put( -2, 22){\rotatebox{90}{\textbf{w/o DT}}}
    \put( 6, 1){\textbf{Reconstruction}}
    \put(35, 1){\textbf{Hausdorff Distance}}
    \put(85, 1){\textbf{SDF}}
    \put(58, 3.5){\textbf{0}}
    \put(55, 30){\textbf{0.0015}}
    \end{overpic}
    \vspace{-3mm}
    \caption{    
    By utilizing a specially designed double-trough~(DT) curve, 
    we allow the presence of non-zero Gaussian curvature, but are more inclined to favor zero Gaussian curvature. Without the usage of the tolerating technique, bulges may arise around the tip points. 
    }
    \label{fig:add_function_results}
\end{figure*}

In Fig.~\ref{fig:add_function_results}, we showcase the effectiveness of the double-trough curve. It can be seen that, in the absence of the tolerating technique, bulges are prone to form around the tip points. Furthermore, with regard to reconstruction accuracy, the application of the double-trough technique results in enhanced accuracy. This improvement is represented through the color-coded visualization of the Hausdorff distance between the reconstructed polygonal surface and the ground truth, underscoring the benefits of the double-trough approach in better representing the SDF of CAD-type models.

\subsection{Implementation Details}
\label{sec:implementation_details}
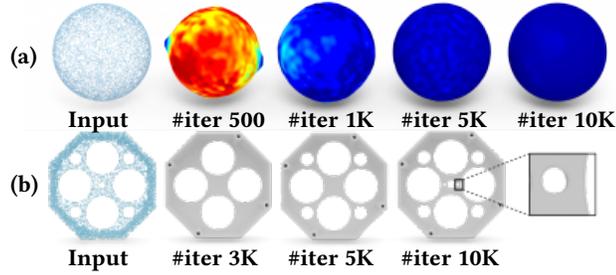
\begin{figure}[t]
    \vspace{2mm}
    \centering
    \begin{overpic}[width=0.95\linewidth]{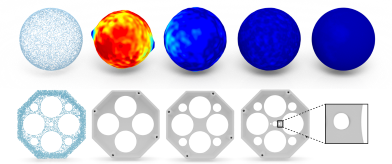}
    \put( -2, 30){\textbf{(a)}}
    \put( 7.5, 19.5){\textbf{Input}}
    \put(24.5, 19.5){\textbf{\#iter 500}}
    \put(43.5, 19.5){\textbf{\#iter 1K}}
    \put(62, 19.5){\textbf{\#iter 5K}}
    \put(81, 19.5){\textbf{\#iter 10K}}
    \put( -2, 9){\textbf{(b)}}
    \put( 7.5, -3){\textbf{Input}}
    \put(24.5, -3){\textbf{\#iter 3K}}
    \put(43.5, -3){\textbf{\#iter 5K}}
    \put(62, -3){\textbf{\#iter 10K}}
    \end{overpic}
    \caption{
    We use an annealing factor to gradually reduce the influence of the Gaussian curvature, 
    such that the fidelity can be preserved even for (a) non-developable surfaces and (b) tiny structures.
    }
    \label{fig:sphere}
    \vspace{-4mm}
\end{figure}

\paragraph{Annealing Factor}
To this end, our total loss is like the following:
\begin{equation}
\label{eq:our_loss}
    \mathcal{L} = \lambda_\text{E} \mathcal{L}_\text{E} + \lambda_\text{DM} \mathcal{L}_\text{DM} + \lambda_\text{DNM} \mathcal{L}_\text{DNM} + \tau\lambda_\text{Gauss} \mathcal{L}_\text{Gauss},
\end{equation}
where~$\tau$ is the annealing factor, which is used to gradually reduce the influence of the Gaussian curvature. 
The annealing process is detailed in Section~\ref{sec:exp_setting}.
In Fig.~\ref{fig:sphere}(a), we illustrate the iterative process using a sphere-shaped point cloud comprising 10K points as input. It is important to note that for a sphere model, the Gaussian curvature cannot be zero for points on the surface. Owing to the implementation of an annealing factor, our Gaussian curvature loss term gradually diminishes throughout the fitting process, ultimately resulting in a high-fidelity representation of the sphere model.
In practice, a CAD model may contain multiple feature parts at significantly different scales. By leveraging the annealing factor, our method demonstrates the capability to reconstruct structures of varying sizes. In Fig.~\ref{fig:sphere}(b), the model includes tiny structures as small as 0.03 in size, within a total size range of $[-0.5, 0.5]^3$. Nonetheless, our algorithm is still able to recover these tiny structures effectively.
\vspace{-1em}

\paragraph{Dynamic Sampling}
In the presence of data imperfections in the given point cloud, such as high sparsity and missing parts, the point set $\Omega$,
around the original point set $\mathcal{P}$,
may prevent the loss term $\mathcal{L}_\text{Gauss}$ from being evaluated in the missing parts.
To address this issue, additional sample points are required near the current surface, enabling $\mathcal{L}_\text{Gauss}$ to more effectively guide the surface's evolution, especially in areas lacking data. 

\begin{figure}[t]
    \vspace{2mm}
    \centering
    \includegraphics[width=0.8\linewidth]{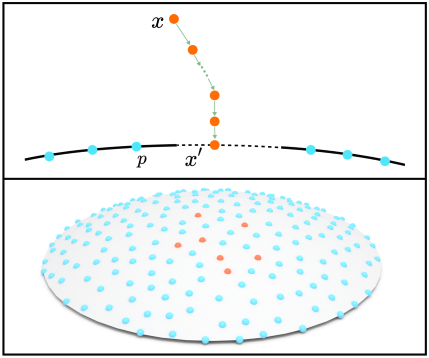}
    \vspace{-1mm}
    \caption{
    We project a point onto the surface based on Eq.~(\ref{eq:proj}).
   The projection points are colored in orange.
    }
    \label{fig:mapping}
    \vspace{-1mm}
\end{figure}

This dynamic sampling process involves projecting each point $\mathcal{Q}$ onto the current surface. For each point $\boldsymbol{x} \in \mathcal{Q}$, the projection $\boldsymbol{x}'$ is calculated as (see Fig.~\ref{fig:mapping}):
\begin{equation}\label{eq:proj}
\boldsymbol{x}' = \boldsymbol{x} - \frac{\nabla f(\boldsymbol{x};\Theta)}{\|\nabla f(\boldsymbol{x};\Theta)\|} \cdot f(\boldsymbol{x};\Theta).
\end{equation}
Subsequently, $\mathcal{L}_\text{Gauss}$ is evaluated for the points in $\Omega \cup \mathcal{Q}'$. As the surface undergoes updates, the set $\mathcal{Q}'$ is updated accordingly.

\begin{figure}[t]
    \centering
     \begin{overpic}[width=0.9\linewidth]{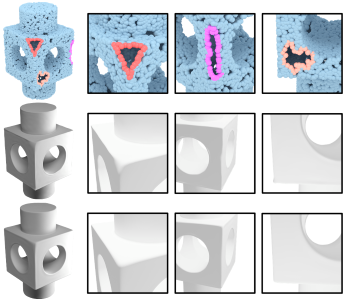}
    \put( -5, 14){{\textbf{(c)}}}
    \put( -5, 42){{\textbf{(b)}}}
    \put( -5, 72){{\textbf{(a)}}}
    \end{overpic}
    \vspace{-2mm}
    \caption{
    In the presence of data imperfections across feature points and lines (a),
    a fixed sampling point set can effectively enforce the Gaussian curve constraints over the surface,
    particularly within the missing parts (b).
    We employ the technique of dynamic sampling to address data imperfections, thereby effectively influencing the surface's evolution, particularly in areas where data may be missing. 
    It is evident that despite the presence of data imperfections across feature points and lines, our algorithm is capable of accurately recovering these actual sharp feature points and lines (c).
    }
    \label{fig:deleteV_recon}
    \vspace{-2mm}
\end{figure}

Without dynamic sampling, 
the Gaussian curvature term cannot be enforced within the data-missing parts. 
In this case, existing approaches fill the missing parts with a smooth transition. 
In this paper, by employing the technique of dynamic sampling to address data imperfections,
we generate additional sample points in close proximity to the current surface, thereby enabling the loss function to more effectively influence the surface's evolution, particularly in areas where data may be missing. As Fig.~\ref{fig:deleteV_recon} shows, it is evident that despite the presence of data imperfections across feature points and lines, our algorithm is capable of accurately recovering these actual sharp feature points and lines.

%% file: sections/04-exp.tex
\section{Experiments}
\label{sec:exp}
In this section, we begin by detailing the implementation specifics and clarifying the metrics employed for evaluation. Following this, a thorough assessment of our NeurCADRecon is carried out across several datasets. Finally, we illustrate its effectiveness in modeling developable surfaces and its competency in handling a range of data imperfections.

\begin{table*}[!t]
\vspace{1.5mm}
\centering
\caption{ Quantitative comparison on the ABC dataset~\cite{ABC}. Each raw point cloud has 10K or 5K points. Methods denoted with a ‘*’ necessitate point normals for their operation, and those marked with ‘$^+$’ rely on supervision-based training. Within each column, the best scores are emphasized with bold and underlining (\underline{\textbf{best}}), whereas the second-best scores are simply highlighted in bold (\textbf{second best}).}
\label{tab:abc}
\resizebox{0.9\textwidth}{!}{
\begin{tabular}{l|cccccc|cccccc} 
\toprule
\multicolumn{1}{c|}{} & \multicolumn{6}{c|}{10K Points}                                                                                                      & \multicolumn{6}{c}{5K Points}                                                                                                 \\ 
\cmidrule{2-13}
\multicolumn{1}{c|}{} & \multicolumn{2}{c}{NC~$\uparrow$} & \multicolumn{2}{c}{CD~$\downarrow$} & \multicolumn{2}{c|}{F1~$\uparrow$} & \multicolumn{2}{c}{NC~$\uparrow$} & \multicolumn{2}{c}{CD~$\downarrow$} & \multicolumn{2}{c}{F1~$\uparrow$}  \\
\multicolumn{1}{c|}{} & mean           & std.                    & mean          & std.                     & mean           & std.                   & mean           & std.                    & mean          & std.                     & mean           & std                    \\ 
\midrule
SPSR$^*$~\footnotesize{\cite{SPSR}}               & 95.16          & 4.48                    & 4.39          & 3.05                     & 74.54          & 26.65                  & 93.06          & 6.31                    & 6.36          & 6.55                     & 62.91          & 30.07                  \\ 
SAL~\footnotesize{\cite{SAL}}                   & 86.25          & 8.39                    & 17.30         & 14.82                    & 29.60          & 18.04                  & 83.85          & 11.28                    & 23.78         & 30.48                     & 28.61          & 18.99                  \\
IGR~\footnotesize{\cite{IGR}}                   & 82.14          & 16.12                   & 36.51         & 40.68                    & 43.47          & 40.06                  & 81.29          & 17.97                   & 37.44         & 41.37                     & 41.14          & 37.57                  \\
SIREN~\footnotesize{\cite{SIREN}}                 & 82.26          & 9.24                    & 17.56         & 15.25                    & 30.95          & 22.23                  & 81.75          & 10.45                    & 18.76         & 19.71                    & 22.02          & 26.92                  \\
Neural-Pull~\footnotesize{\cite{NeuralPull}}           & 94.23          & 4.57                    & 6.73          & 5.15                     & 42.67          & \under{10.75}         & 92.52          & 8.51                    & 6.97          & 10.37            & 72.32          & 27.74          \\
DiGS~\footnotesize{\cite{DiGS}}                 & 94.48          & 6.12                    & 6.91          & 6.94                     & 66.22          & 32.01                  & 92.55          & 7.11                    & 7.91          & 6.96                     & 55.86          & 31.52                  \\
iPSR~\footnotesize{\cite{iPSR}}                  & 93.15          & 7.47                    & 4.84          & 4.06                     & 71.59          & 24.96                  & 90.57          & 8.71                    & 6.21          & 4.49                     & 61.27          & 26.23                  \\
PGR~\footnotesize{\cite{PGR}}                   & 94.11          & 4.63                    & 4.52          & 2.13                     & 68.91          & 27.86                  & 89.29          & 7.91                    & 7.96          & 3.76                     & 43.27          & 23.12                  \\ 
POCO$^+$~\footnotesize{\cite{POCO}}              & 92.90          & 7.00                    & 6.05          & 6.80                     & 68.29          & 26.05                  & 90.24          & 5.71                    & 7.93          & 5.07                     & 48.25          & 23.89                  \\
NG$^+$~\footnotesize{\cite{GalerkinNG}}   & 95.88      &  3.88              &  3.60                       &      \under{1.38}         &   81.38                & 20.39               &   94.07                    & 5.31              &   4.05                      &  1.71            &  76.42                        &      21.03                                 \\ 
NSH~\footnotesize{\cite{HessianZX}}   & \textbf{97.42}       &  \textbf{2.37}              &  \textbf{3.27}                       &      1.78         &    \textbf{88.62}                & 13.87               &   \textbf{94.81}                     & \textbf{3.57}               &   \textbf{3.97}                      &  \under{1.18}             &  \textbf{78.86}                        &     \textbf{17.15}                                 \\ 
\midrule
\textbf{Ours}                  & \under{97.57} & \under{2.01}           & \under{3.12} & \textbf{1.41}            & \under{89.03} & \textbf{13.05}                  & \under{96.29} & \under{3.55}           & \under{3.47} & \textbf{1.46}                     & \under{85.34} & \under{15.17}                  \\
\bottomrule
\end{tabular}
}
\vspace{1.5mm}
\end{table*}

\subsection{Experimental Setting}
\label{sec:exp_setting}

\paragraph{Parameters and Platform}
Similar to various implicit surface reconstruction methods~\cite{IDF, phase, DiGS}, our NeurCADRecon employs the SIREN~\cite{SIREN} network architecture, featuring four hidden layers, each with 256 units. The SIREN~\cite{SIREN} architecture comprises layers based on MLPs, with inputs initially normalized to the range $[-1, 1]^3$ before being fed into the network. The activation function employed in this method is the sine periodic function. In our experiments, we determine the weights $(\lambda_\text{E}=50, \lambda_{\text{DM}}=7000, \lambda_{\text{DNM}}=600, \lambda_{\text{Gauss}}=10)$ based on our tailored configurations. The annealing factor $\tau$ remains at 1 during the initial 20\% of iterations, then linearly decreases to $1 \times 10^{-4}$ from~20\% to 50\% of the iteration span, and finally drops to 0 towards the end. We extract the zero-level set of the implicit function into a discrete mesh using the marching cubes algorithm \cite{lewiner2003efficient} with a $256^3$ grid. Throughout the training phase, we apply the Adam optimizer \cite{Adam} with a default learning rate of~$5 \times 10^{-5}$ and complete 10K iterations. The experiments detailed in this paper were executed on an NVIDIA GeForce RTX 3090 graphics card equipped with 24GB of video memory and powered by an AMD EPYC 7642 processor.

\paragraph{Evaluation Metrics}
To evaluate the accuracy of the reconstructed mesh, we utilize three primary metrics: Chamfer Distance (CD), F1-score (F1), and Normal Consistency (NC). CD, scaled by $10^3$ and calculated using the $L1$-norm, quantifies the similarity between two surfaces. F1, scaled by $10^2$ and employing a default threshold of~$5\times 10^{-3}$, represents the harmonic mean of precision and recall, assessing the balance between the detail capture and the inclusion of extraneous elements. NC, also scaled by $10^2$, measures the agreement between the normals of the reconstructed surface and those of the ground-truth, indicating the directional accuracy of the surface details.
The mean values along with the standard deviation of the reconstruction outcomes across each dataset are documented to provide a comprehensive overview of the performance.

\paragraph{Datasets}
We carry out overfitting surface reconstruction experiments on four recent CAD datasets: ABC~\cite{ABC}, Fusion Gallery~\cite{DatasetFG}, DeepCAD~\cite{DatasetDeepCAD}, and the dataset utilized by CAPRI-Net~\cite{CAPRI_Net}. To maintain uniformity in evaluation, all meshes are scaled to fit within the range of $[-0.5, 0.5]^3$, and 10K points are sampled from each mesh, ensuring a consistent and fair basis for comparison across all datasets.

\subsection{Comparison on Open Datasets}
\label{sec:compare_dataset}

We evaluate our proposed method, NeurCADRecon, against three categories of state-of-the-art surface reconstruction techniques:
\begin{itemize}
\item Approaches with Normals: In this category, we include the classic surface reconstruction method SPSR~\cite{SPSR} for comparison, which utilizes normals in its process.
\item Supervision-based Approaches: This category comprises two methods, POCO~\cite{POCO} and Neural Galerkin (NG)~\cite{GalerkinNG}, both of which depend on ground truth data for training. To enhance the performance of POCO and NG, we retrained these methods on the comprehensive ShapeNet dataset~\cite{ShapeNet} using 10K input points prior to comparing them with our approach.

\item Overfitting-based Approaches: Methods in this category leverage various reconstruction loss constraints for self-supervised learning, including DiGS~\cite{DiGS}, SIREN~\cite{SIREN}, Neural-Pull~\cite{NeuralPull}, IGR~\cite{IGR}, SAL~\cite{SAL}, CAPRI-Net~\cite{CAPRI_Net}, iPSR~\cite{iPSR}, NSH~\cite{HessianZX}, and our own NeurCADRecon. It's noteworthy that CAPRI-Net~\cite{CAPRI_Net} operates in two phases: pre-training and fine-tuning, with a detailed comparison provided in Subsection~\ref{sec:abc_all}.
\end{itemize}

\begin{figure*}[!t]
    \centering
    \vspace{1mm}
    \begin{overpic}[width=0.98\linewidth]{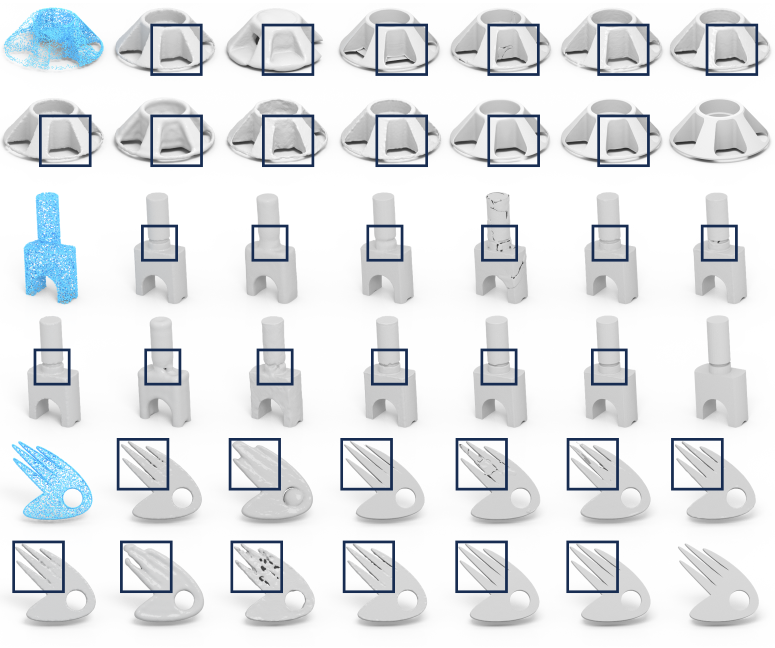}
\put( 5,14.5){\textbf{Input}}
\put(19,14.5){\textbf{SPSR$^*$}}
\put(34,14.5){\textbf{SAL}}
\put(49,14.5){\textbf{IGR}}
\put(62,14.5){\textbf{SIREN}}
\put(75,14.5){\textbf{Neural-Pull}}
\put(91,14.5){\textbf{DiGS}}
\put(5,0.5){\textbf{iPSR}}
\put(20,0.5){\textbf{PGR}}
\put(33,0.5){\textbf{POCO$^+$}}
\put(49,0.5){\textbf{NG$^+$}}
\put(63,0.5){\textbf{NSH}}
\put(77,0.5){\textbf{Ours}}
\put(92,0.5){\textbf{GT}}

\put( 5,43.1){\textbf{Input}}
\put(19,43.1){\textbf{SPSR$^*$}}
\put(34,43.1){\textbf{SAL}}
\put(49,43.1){\textbf{IGR}}
\put(62,43.1){\textbf{SIREN}}
\put(75,43.1){\textbf{Neural-Pull}}
\put(91,43.1){\textbf{DiGS}}
\put( 5,27.4){\textbf{iPSR}}
\put(20,27.4){\textbf{PGR}}
\put(33,27.4){\textbf{POCO$^+$}}
\put(49,27.4){\textbf{NG$^+$}}
\put(63,27.4){\textbf{NSH}}
\put(77,27.4){\textbf{Ours}}
\put(92,27.4){\textbf{GT}}

\put( 5,72){\textbf{Input}}
\put(19,72){\textbf{SPSR$^*$}}
\put(34,72){\textbf{SAL}}
\put(49,72){\textbf{IGR}}
\put(62,72){\textbf{SIREN}}
\put(75,72){\textbf{Neural-Pull}}
\put(91,72){\textbf{DiGS}}
\put( 5,60){\textbf{iPSR}}
\put(20,60){\textbf{PGR}}
\put(33,60){\textbf{POCO$^+$}}
\put(49,60){\textbf{NG$^+$}}
\put(63,60){\textbf{NSH}}
\put(77,60){\textbf{Ours}}
\put(92,60){\textbf{GT}}
    \end{overpic}
    \vspace{-4mm}
    \caption{
    Comparison with state-of-the-art surface reconstruction methods is conducted on a variety of challenging CAD models, including those with thin tubes, tiny links, sharp corners, and narrow slits. The first model showcases reconstruction results of varying density input points (10K and 5K simultaneously), while the remaining two models are based on 10K input points. Method marked with `$^*$' requires normals, and those marked with `$^+$' are supervision-based. Across the board, our NeurCADRecon outperforms other methods in terms of reconstruction fidelity on complex CAD models.
    }
    \label{fig:comp_CAD10k}
    \vspace{-2mm}
\end{figure*}

In Fig.~\ref{fig:comp_CAD10k}, we showcase a variety of challenging CAD models, such as thin tubes, tiny links, sharp corners, and narrow slits, to qualitatively demonstrate that our results are capable of handling complex CAD geometries in comparison with the aforementioned methods.

\begin{table}[!t]
\vspace{1.5mm}
\centering
\caption{
 Quantitative comparison on the Fusion Gallery dataset~\cite{DatasetFG}.
 Each raw point cloud has 10K points.
}
\label{tab:FG}
 \resizebox{.48\textwidth}{!}{%
\begin{tabular}{l|cccccc} 
\toprule
\multicolumn{1}{c|}{} & \multicolumn{2}{c}{NC~$\uparrow$} & \multicolumn{2}{c}{CD~$\downarrow$} & \multicolumn{2}{c}{F1~$\uparrow$} \\
\multicolumn{1}{c|}{} & mean           & std.                    & mean          & std.                     & mean           & std. \\ 
\midrule
SPSR$^*$~\footnotesize{\cite{SPSR}}               & 97.33          & 3.35                    & 3.01          & 2.48                     & 88.51          & 18.61                \\ 
SAL~\footnotesize{\cite{SAL}}                   & 92.42          & 7.58                    & 13.31         & 21.98                    & 47.16          & 28.45                  \\
IGR~\footnotesize{\cite{IGR}}                   & 87.07          & 14.71                   & 29.75         & 37.73                    & 54.09          & 42.67            \\
SIREN~\footnotesize{\cite{SIREN}}                 & 95.31          & 5.22                    & 6.38         & 13.45                    & 80.85          & 22.28                              \\
Neural-Pull~\footnotesize{\cite{NeuralPull}}           & 98.05          & 3.25                    & 3.93          & 7.26                     & 89.99          & 20.51\\
DiGS~\footnotesize{\cite{DiGS}}                 & 97.53          & 3.01                    & 4.96          & 6.62                     & 80.77          & 29.28  \\
iPSR~\footnotesize{\cite{iPSR}}                  & 96.05          & 7.36                    & 3.16          & 2.65                     & 86.48          & 19.48                 \\
PGR~\footnotesize{\cite{PGR}}                   & 93.55          & 8.19                    & 4.84          & 3.13                     & 67.52          & 26.36                 \\ 
POCO$^+$~\footnotesize{\cite{POCO}}              & 94.63          & 3.45                    & 6.91          & 7.04                     & 69.31          & 28.08            \\
NG$^+$~\footnotesize{\cite{GalerkinNG}}   & 97.87       &  \textbf{2.79}              &  \under{2.68}                       &      \under{1.29}         &    90.78                & 15.11                                            \\ 
NSH~\footnotesize{\cite{HessianZX}}   & \under{98.99}       &  4.79              &  3.12                       &      1.71         &    \textbf{90.92}                & \textbf{11.25}               \\ 
\midrule
\textbf{Ours}                  & \textbf{98.07} & \under{2.39}           & \textbf{2.75} & \textbf{1.69}            & \under{92.99} & \under{10.01}                \\
\bottomrule
\end{tabular}
}
\end{table}

\begin{table}[!t]
\vspace{1.5mm}
\centering
\caption{
 Quantitative comparison on the DeepCAD dataset~\cite{DatasetDeepCAD}.
 Each raw point cloud has 10K points.
}
\label{tab:deepcad}
 \resizebox{.48\textwidth}{!}{%
\begin{tabular}{l|cccccc} 
\toprule
\multicolumn{1}{c|}{} & \multicolumn{2}{c}{NC~$\uparrow$} & \multicolumn{2}{c}{CD~$\downarrow$} & \multicolumn{2}{c}{F1~$\uparrow$} \\
\multicolumn{1}{c|}{} & mean           & std.                    & mean          & std.                     & mean           & std. \\ 
\midrule
SPSR$^*$~\footnotesize{\cite{SPSR}}               & 97.73          & 2.47                    & 3.12          & 3.91                     & 87.43          & 19.74                \\ 
SAL~\footnotesize{\cite{SAL}}                   & 93.64          & 7.11                    & 11.48         & 17.27                    & 45.64          & 27.83                  \\
IGR~\footnotesize{\cite{IGR}}                   & 89.73          & 13.26                   & 27.18         & 37.25                    & 59.99          & 41.57            \\
SIREN~\footnotesize{\cite{SIREN}}                 & 94.77          & 6.46                    & 8.79         & 20.27                    & 76.16          & 25.91                              \\
Neural-Pull~\footnotesize{\cite{NeuralPull}}           & 97.41          & 4.36                    & 4.81          & 11.34                     & 86.87          & 21.64\\
DiGS~\footnotesize{\cite{DiGS}}                 & 97.74          & 2.39                    & 5.24          & 6.84                     & 77.93          & 31.12  \\
iPSR~\footnotesize{\cite{iPSR}}                  & 96.99          & 4.84                    & 3.25          & 2.66                     & 85.29          & 20.25                 \\
PGR~\footnotesize{\cite{PGR}}                   & 94.85          & 5.51                    & 4.73          & 2.66                     & 67.26          & 25.91                 \\ 
POCO$^+$~\footnotesize{\cite{POCO}}              & 94.62          & 2.89                    & 8.48          & 8.41                     & 65.66          & 28.86            \\
NG$^+$~\footnotesize{\cite{GalerkinNG}}   & \under{98.18}       &  \under{2.12}              &  \under{3.01}                       &      \under{2.12}         &    \textbf{88.36}                & \textbf{17.17}                                           \\
NSH~\footnotesize{\cite{HessianZX}}   & 97.98       &  2.38              &  3.15                       &      2.39         &    87.63                & 24.16                                \\ 
\midrule
\textbf{Ours}                  & \textbf{98.03} & \textbf{2.17}           & \textbf{3.08} & \textbf{2.26}            & \under{90.70} & \under{12.82}                \\
\bottomrule
\end{tabular}
}
\end{table}

\paragraph{ABC Dataset}
The ABC dataset, as described in~\citet{ABC}, comprises a diverse assortment of CAD meshes. 
Following the approach of~\citet{Points2surf}, we select 100~\cite{Points2surf, NeuralPull, HessianZX} mechanically intricate yet clean and watertight shapes from the dataset and randomly sample 10K or 5K points from each mesh.
In Tab.~\ref{tab:abc}, quantitative comparison statistics between our method and the baseline approaches are presented.
Our method demonstrates superior performance across all three evaluation metrics, whether applied to input point clouds with 10K or 5K points. Notably, when reconstructing CAD models with only 5K input points, our F1 surpasses that of the second-best method by 6.48\%.
Then, we show the results of the reconstruction using varying points~(5K and 10K) on one model (see the first model in Fig.~\ref{fig:comp_CAD10k}).
In visualization comparison with NSH~\cite{HessianZX}, our method successfully reconstructs the thin tube structure in its entirety. In addition, our reconstruction results do not have redundant faces like the other methods.

\vspace{-2mm}
\paragraph{Fusion Gallery Dataset}
The Fusion Gallery dataset, as described in~\citet{DatasetFG}, consists of 8625 human-designed CAD models generated using profile sketches and extrusions, with all CAD models derived from real-world application scenarios.
In the analysis presented in this paper, we randomly selected 10K points from each mesh and input them to all approaches for comparison.
Quantitative comparison statistics are provided in Tab.~\ref{tab:FG}.
The statistics indicate that our method consistently achieves optimal results on average, demonstrating that the CAD models reconstructed by our method closely align with those designed by humans.
In the comparison with tiny links (the second model in Fig.\ref{fig:comp_CAD10k}), SAL~\cite{SAL}, PGR~\cite{PGR}, and POCO~\cite{POCO} lose the part. Other methods create uneven surfaces, leading to the loss of feature lines. In contrast, our method successfully reconstructs this structure while maintaining the feature line.
\vspace{-2mm}

\paragraph{DeepCAD Dataset} 
The DeepCAD dataset, detailed in~\citet{DatasetDeepCAD}, comprises 178,238 CAD models created via sketch-extrude operations. In contrast to the Fusion Gallery Dataset~\cite{DatasetFG}, this dataset contains more CAD models with sharp features, such as sharp corners. In Tab.~\ref{tab:deepcad}, we present the reconstruction results on the DeepCAD dataset, with all methods randomly sampling 10K points as input.
Quantitatively, our method performs consistently with NG~\cite{GalerkinNG} on NC and CD, but we outperform it by 1.8\% on F1. The visual comparison results of sharp corners and narrow slits can be observed in the third model of Fig.~\ref{fig:comp_CAD10k}. From the visualization, although many methods can reconstruct sharp corners, only our method and NSH~\cite{HessianZX} accurately reconstruct narrow slits close to the ground truth.

\vspace{-1mm}
\paragraph{CAD Dataset from CAPRI-Net}
\label{sec:abc_all}
The dataset utilized by CAPRI-Net~\cite{CAPRI_Net} includes CAD models from ABC~\cite{ABC}, where each model consists of complex components (as shown in Fig.\ref{fig:gallery}) and is provided in both point cloud and voxel forms. In this comparison, we input voxels to CAPRI-Net\cite{CAPRI_Net}, while other methods take a point cloud with randomly sampled 10K points as input.
Tab.~\ref{tab:abc_all} presents the quantitative comparison results between our method and other methods, excluding CAPRI-Net~\cite{CAPRI_Net}. We achieved optimal results across all three evaluation metrics.

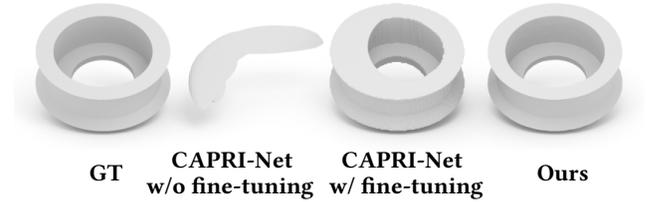
\begin{figure}[t]
    \vspace{-1.5mm}
    \centering
\begin{overpic}[width=0.98\linewidth]{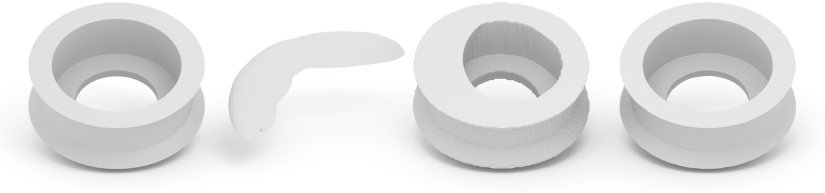}
    \put( 12,-5){\textbf{GT}}
    \put(25,-3){\textbf{CAPRI-Net}}
    \put(21,-7){\textbf{w/o fine-tuning}}
    \put(52,-3){\textbf{CAPRI-Net}}
    \put(50,-7){\textbf{w/ fine-tuning}}
    \put(83,-5){\textbf{Ours}}
    \end{overpic}
    \vspace{5mm}
    \caption{
    Visualizing the reconstruction results between CAPRI-Net~\cite{CAPRI_Net} and our NeurCADRecon.
    }
    \label{fig:compare_capriNet}
\end{figure}

To perform a comprehensive comparison with CAPRI-Net~\cite{CAPRI_Net}, we utilized the publicly available data and network weights of this method without re-training. Tab.~\ref{tab:compare_capriNet} lists the quantitative results of CAPRI-Net~\cite{CAPRI_Net} in both with and without fine-tuning phase, and Fig.~\ref{fig:compare_capriNet} displays the corresponding reconstruction results.
Our method consistently outperforms CAPRI-Net~\cite{CAPRI_Net} in both quantitative and qualitative aspects. It is worth noting that CAPRI-Net~\cite{CAPRI_Net} may encounter alignment issues with different predicted primitives during Boolean combination operations, leading to reconstruction results with redundant faces.

\begin{table}[!t]
\vspace{1.5mm}
\centering
\caption{
 Quantitative comparison on the CAPRI-Net~\cite{CAPRI_Net} dataset.
 Each raw point cloud has 10K points.
}
\label{tab:abc_all}
 \resizebox{.48\textwidth}{!}{%
\begin{tabular}{l|cccccc} 
\toprule
\multicolumn{1}{c|}{} & \multicolumn{2}{c}{NC~$\uparrow$} & \multicolumn{2}{c}{CD~$\downarrow$} & \multicolumn{2}{c}{F1~$\uparrow$} \\
\multicolumn{1}{c|}{} & mean           & std.                    & mean          & std.                     & mean           & std. \\ 
\midrule
SPSR$^*$~\footnotesize{\cite{SPSR}}               & 97.56          & 1.41                    & 3.11          & 1.39                     & 86.59          & 17.09                \\ 
SAL~\footnotesize{\cite{SAL}}                   & 92.71          & 5.38                    & 10.88         & 8.52                    & 44.29          & 22.43                 \\
IGR~\footnotesize{\cite{IGR}}                   & 91.36          & 12.22                   & 10.55         & 14.12                    & 67.07          & 30.93            \\
SIREN~\footnotesize{\cite{SIREN}}                 & 95.07          & 5.59                    & 8.28         & 18.28                    & 76.27          & 24.58                              \\
Neural-Pull~\footnotesize{\cite{NeuralPull}}           & 97.02          & 5.89                    & 6.23          & 18.02                     & 86.44          & 21.83\\
DiGS~\footnotesize{\cite{DiGS}}                 & 96.43          & 4.24                    & 7.44          & 8.93                     & 70.29          & 35.43  \\
iPSR~\footnotesize{\cite{iPSR}}                  & 96.97          & 2.36                    & 3.34          & 1.69                     & 83.86          & 18.37                 \\
PGR~\footnotesize{\cite{PGR}}                   & 94.26          & 3.77                    & 5.18          & 2.07                     & 60.37          & 22.76                 \\ 
POCO$^+$~\footnotesize{\cite{POCO}}              & 93.76          & 3.36                    & 7.07          & 5.72                     & 61.18          & 26.22            \\
NG$^+$~\footnotesize{\cite{GalerkinNG}}   & \textbf{98.15}       &  \textbf{1.34}              &  \textbf{2.93}                       &      \under{1.06}         &    \textbf{89.01}                & \textbf{15.01}                                           \\ 
NSH~\footnotesize{\cite{HessianZX}}   & 97.48      &  1.47              &  4.21                       &      3.88         &    85.81                & 15.53                                 \\ 
\midrule
\textbf{Ours}                  & \under{98.42} & \under{1.10}           & \under{2.84} & \textbf{1.36}            & \under{92.79} & \under{9.64}                \\
\bottomrule
\end{tabular}
}
\end{table}

\subsection{Imperfection Data and Further Comparison}
\begin{figure*}[t]
    \centering
    \vspace{1.5mm}
    \begin{overpic}[width=0.98\linewidth]{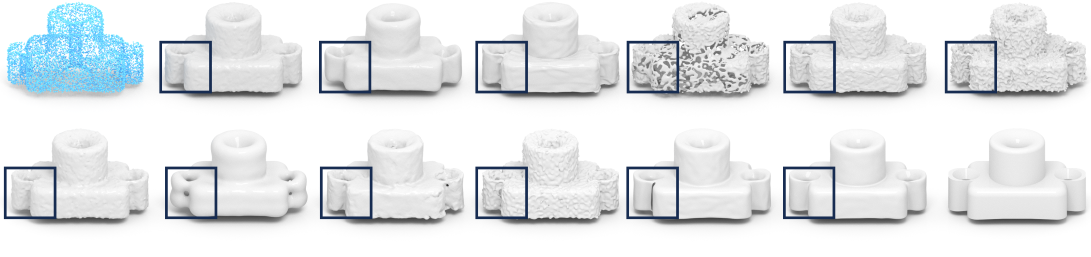}
\put( 5,12.7){\textbf{Input}}
\put(19,12.7){\textbf{SPSR$^*$}}
\put(34,12.7){\textbf{SAL}}
\put(49,12.7){\textbf{IGR}}
\put(62,12.7){\textbf{SIREN}}
\put(74,12.7){\textbf{Neural-Pull}}
\put(91,12.7){\textbf{DiGS}}
\put( 5,1){\textbf{iPSR}}
\put(20,1){\textbf{PGR}}
\put(33,1){\textbf{POCO$^+$}}
\put(49,1){\textbf{NG$^+$}}
\put(63,1){\textbf{NSH}}
\put(77,1){\textbf{Ours}}
\put(92,1){\textbf{GT}}
    \end{overpic}
     \vspace{-4mm}
    \caption{
   Evaluate various point surface reconstruction approaches on a point cloud with 0.5\% Gaussian noise. Our method stands out as the only approach that effectively recovers a smooth CAD surface even in the presence of input points affected by noise.
    }
    \label{fig:comp_noise}
\end{figure*}

\paragraph{Noise}
\label{sec:noise}
To evaluate the robustness of our method against noise, we introduced point clouds with reconstruction noise. We specifically used a 10K point cloud as the baseline input and added 0.5\% Gaussian noise to test the noise immunity of our network. 
For a comprehensive comparison, we input the same point cloud to the other 11 methods under identical noise conditions. As depicted in Fig.~\ref{fig:comp_noise}, our approach outperforms the alternative methods in the noisy scenarios, which validates that our method can reconstruct faithful CAD models with smooth surfaces.
The scalability of our method plays a significant role in its superior smoothing performance, which is instrumental in its effective resistance to noise. A more detailed comparison of our method with other smoothing functions is provided in Section~\ref{sec:smooth_fun}.

\begin{figure*}[!t]
    \centering
    \vspace{-2mm}
    \begin{overpic}[width=0.98\linewidth]{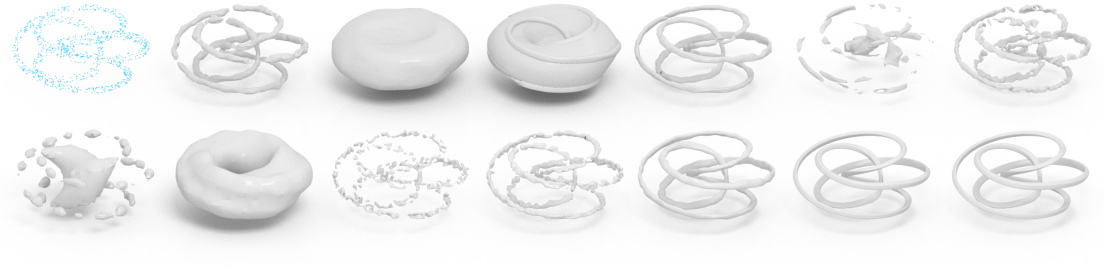}
\put( 5,13){\textbf{Input}}
\put(19,13){\textbf{SPSR$^*$}}
\put(34,13){\textbf{SAL}}
\put(49,13){\textbf{IGR}}
\put(61,13){\textbf{SIREN}}
\put(74,13){\textbf{Neural-Pull}}
\put(91,13){\textbf{DiGS}}
\put( 5,1.5){\textbf{iPSR}}
\put(20,1.5){\textbf{PGR}}
\put(33,1.5){\textbf{POCO$^+$}}
\put(49,1.5){\textbf{NG$^+$}}
\put(62,1.5){\textbf{NSH}}
\put(76,1.5){\textbf{Ours}}
\put(91,1.5){\textbf{GT}}
    \end{overpic}
    \vspace{-4mm} 
    \caption{
    Experiments are conducted on sparse point clouds containing only 1K points, and our results exhibit a close alignment with the ground truth for the provided input. This comparison highlights the substantial advantage of our algorithm in dealing with sparse raw data.
    }
    \label{fig:comp_sparse}
\end{figure*}

\paragraph{Data Sparsity}
In Fig.~\ref{fig:comp_sparse}, we present an example with a sparse point cloud containing 1K points. This specific case is designed to assess the algorithm's performance on sparse inputs, as nearby gaps and thin-walled tubes/plates can significantly increase the difficulty of predicting SDFs in situations of data sparsity. The visual comparison highlights that only our method, SIREN~\cite{SIREN}, and NSH~\cite{HessianZX} successfully reconstruct the geometric structure in accordance with the input, without exhibiting undesired genus. Furthermore, compared to these two methods, our reconstruction results demonstrate smoother and closer alignment with the ground truth.

\begin{figure*}[!t]
    \vspace{2mm}
    \centering
    \begin{overpic}[width=0.98\linewidth]{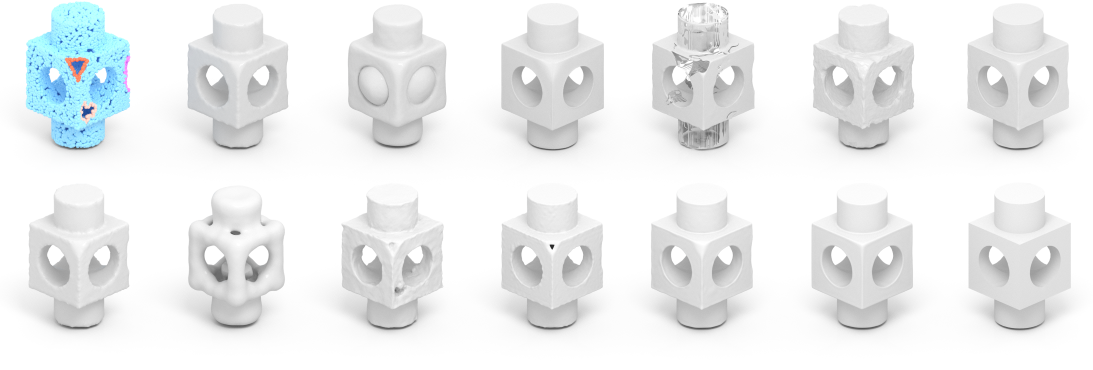}
\put( 4,18){\textbf{Input}}
\put(19,18){\textbf{SPSR$^*$}}
\put(34.5,18){\textbf{SAL}}
\put(48.5,18){\textbf{IGR}}
\put(62,18){\textbf{SIREN}}
\put(74.5,18){\textbf{Neural-Pull}}
\put(91.5,18){\textbf{DiGS}}
\put( 5,1.5){\textbf{iPSR}}
\put(19,1.5){\textbf{PGR}}
\put(33,1.5){\textbf{POCO$^+$}}
\put(49,1.5){\textbf{NG$^+$}}
\put(62.5,1.5){\textbf{NSH}}
\put(77,1.5){\textbf{Ours}}
\put(92,1.5){\textbf{GT}}
    \end{overpic}
    \vspace{-4mm}
    \caption{
    In a comparison with state-of-the-art methods using an incomplete point cloud as input, characterized by data imperfections across feature points and lines, our NeurCADRecon outperforms other methods in terms of reconstruction accuracy.
    }
    \label{fig:comp_missing}
\end{figure*}

\paragraph{Incomplete Point Cloud}
The presence of an incomplete point cloud poses a challenge for reconstructing from an unoriented point cloud. In Fig.~\ref{fig:comp_missing}, we present a model with missing points across feature points and lines of the input points. 
The observations demonstrate that, with the exception of our method, other approaches encounter difficulties in reconstructing the expected sharp features at locations where points are missing. In such cases, existing methods tend to produce smooth transition surfaces. However, our approach is unaffected by this issue due to our dynamic sampling strategy, which introduces supplementary sampling points in proximity to the current surface. This enhances the influence of the loss function on surface optimization, resulting in a reconstruction more aligned with our loss function settings. Consequently, our method excels in reconstructing sharp feature points and lines even in the absence of certain sampling points.

\begin{table}[!tp]
\vspace{1.5mm}
\centering
\caption{Comparing our approach with CAPRI-Net~\cite{CAPRI_Net}.
}
\label{tab:compare_capriNet}
 \resizebox{\linewidth}{!}{%
\begin{tabular}{l|c|cccccc} 
\toprule
\multirow{2}{*}{}     & w/ & \multicolumn{2}{c}{NC~$\uparrow$}  & \multicolumn{2}{c}{CD~$\downarrow$} & \multicolumn{2}{c}{F1~$\uparrow$}  \\
  &    fine-tuning       & mean           & std.          & mean          & std.                     & mean           & std.                  \\ 
\midrule
CAPRI-Net~\cite{CAPRI_Net}  &  $\times$
& 61.79 & 13.76 & 92.96 & 59.29 & 5.86 & \under{8.41} \\
CAPRI-Net~\cite{CAPRI_Net}  & \checkmark
& 92.06   & 3.82 & 9.42 & 3.16 & 34.47 & 18.36 \\
\midrule
\textbf{Ours} &  $\times$
& \under{98.42} & \under{1.10}           & \under{2.84} & \under{1.36}            & \under{92.79} & 9.64  \\
\bottomrule
\end{tabular}
}
\end{table}

\begin{figure}[t]
    \vspace{3mm}
    \centering
\begin{overpic}[width=0.98\linewidth]{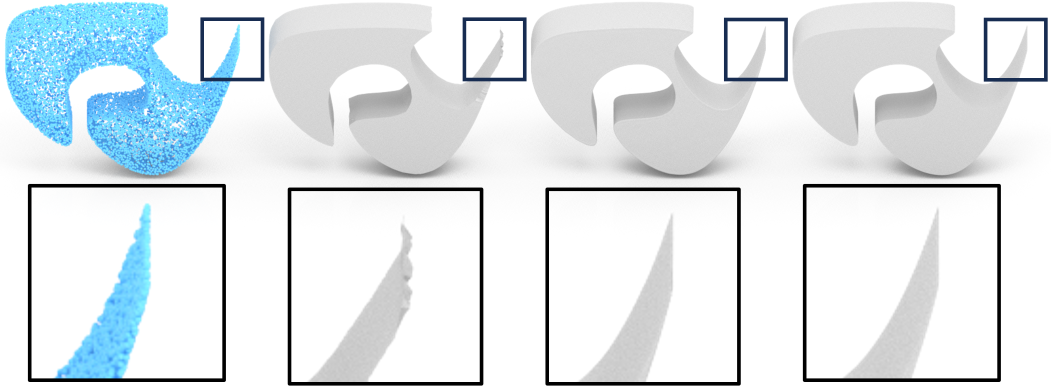}
\put( 8,-4){\textbf{Input}}
\put(33,-4){\textbf{RFEPS}}
\put(59,-4){\textbf{Ours}}
\put(84,-4){\textbf{GT}}
    \end{overpic}
    \vspace{1mm}
    \caption{
    Visualizing the reconstruction results between RFEPS~\cite{Rui2022RFEPS} and our NeurCADRecon.
    }
    \label{fig:compare_RFEPS}
\end{figure}

\paragraph{Comparison with RFEPS}
In a recent seminal work on explicit reconstruction, RFEPS~\cite{Rui2022RFEPS} introduces a method for reconstructing a polygonal surface that incorporates feature lines, utilizing the concept of the restricted power diagram.
However, due to the constraints of the restricted power diagram, RFEPS~\cite{Rui2022RFEPS} faces challenges in reconstructing positions within the CAD model where the dihedral angle is less than $\pi/3$. In contrast, our method is free from this limitation and successfully reconstructs sharp features, as illustrated in Fig.~\ref{fig:compare_RFEPS}. Additionally, RFEPS~\cite{Rui2022RFEPS} incurs significant time overhead, requiring 3630 milliseconds for a 10K point cloud test, while our method achieves the same task in only 39.58 milliseconds, demonstrating a 92 times improvement in time efficiency.

\begin{figure}[t]
    \vspace{3mm}
    \centering
    \begin{overpic}[width=0.95\linewidth]{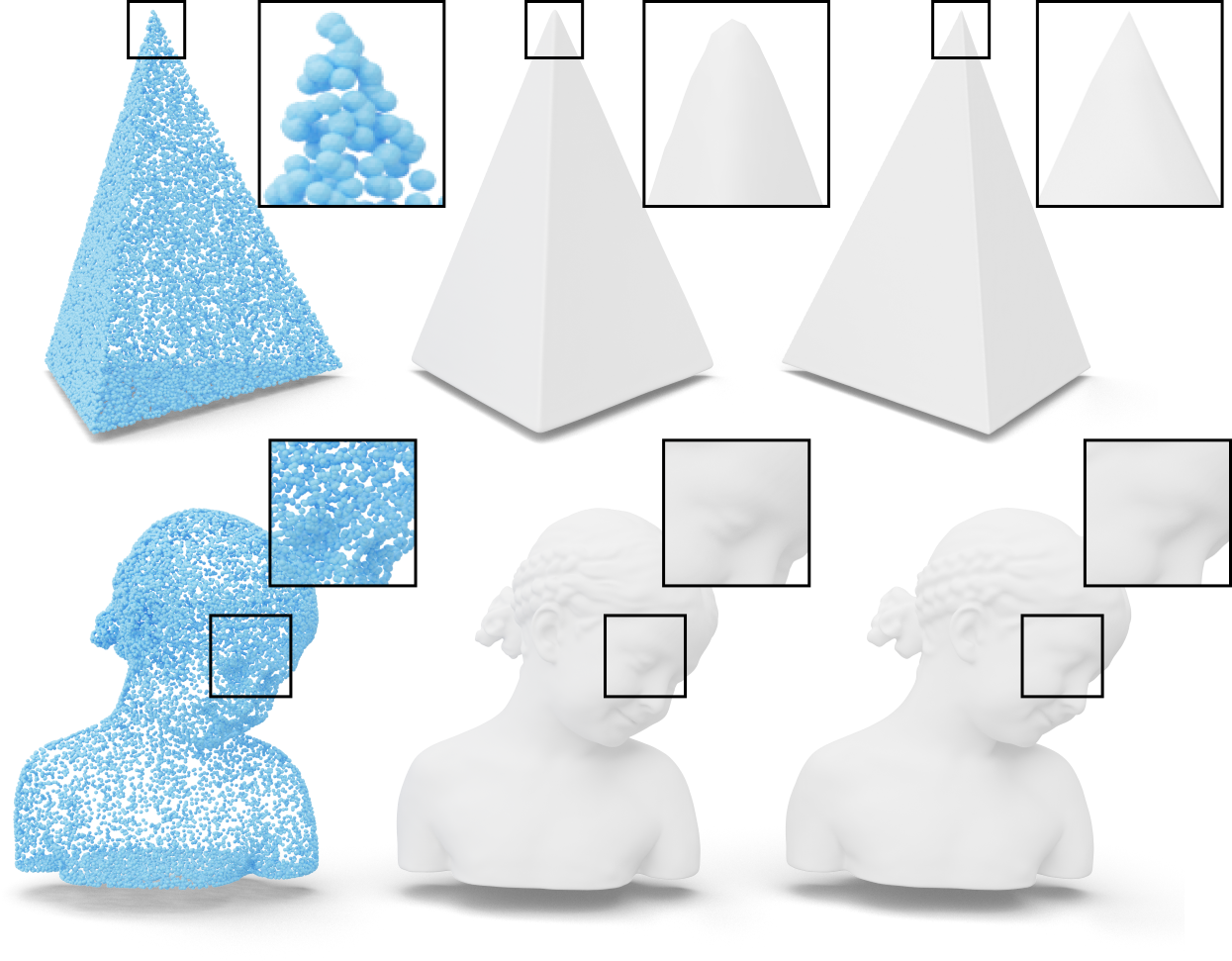}
    \put( -2,55){\textbf{(a)}}
    \put( -2,18){\textbf{(b)}}
    \put( 12,-2){\textbf{Input}}
    \put(43,-2){\textbf{NSH}}
    \put(75,-2){\textbf{Ours}}
    \end{overpic}
    \caption{
    Visualizing the reconstruction results between NSH~\cite{HessianZX} and our NeurCADRecon.
    }
    \label{fig:compare_NSH}
\end{figure}

\paragraph{Comparison with NSH}
As previously mentioned, NSH~\cite{HessianZX} encourages the Hessian matrix to be singular, aiming to eliminate ghost geometry while preserving geometric details. This approach is related to, but distinct from our method. Generally, for points near a surface, the Hessian of the SDF aligns with the non-vanishing gradients, with an eigenvalue of 0, rendering the Hessian singular. Only a very limited number of singular points may exist where the Hessian has a non-zero determinant, such as Morse saddle points of the SDF. Thus, NSH~\cite{HessianZX} aims to reduce the number of such singularities without introducing extra smoothness. This is why NSH~\cite{HessianZX} excels at preserving details.

Our method, on the other hand, encourages every point to be a planar point (including feature-line points), except for a limited number of corner points (where~$k_{\text{Gauss}}$ is significantly different from 0). The introduction of the double-trough function is designed to force the Gaussian curvature to follow a polarization distribution. NeurCADRecon is more effective in recovering sharp feature lines (as seen in the sharp corner in Fig.~\ref{fig:compare_NSH}(a)), while NSH is better at reconstructing shapes with rich details (as seen in the eye of the free-form model in Fig.~\ref{fig:compare_NSH}(b)). To summarize, NeurCADRecon is tailored for CAD models, whereas NSH~\cite{HessianZX} is intended for non-CAD models.

Additionally, our regularization accommodates the existence of non-90-degree dihedral angles. Although the model in Fig.~\ref{fig:compare_NSH}(a) features a non-90-degree dihedral angle, our method can still accurately reconstruct non-90-degree corners. Fundamentally, our double-trough function encourages~$k_{\text{Gauss}}$ to be 0 or~$\pi/2$, but the entire~$k_{\text{Gauss}}$ distribution must conform to the Gauss-Bonnet theorem~\cite{GaussBonnet}.

\begin{figure}[t]
    \vspace{3mm}
    \centering
\begin{overpic}[width=0.95\linewidth]{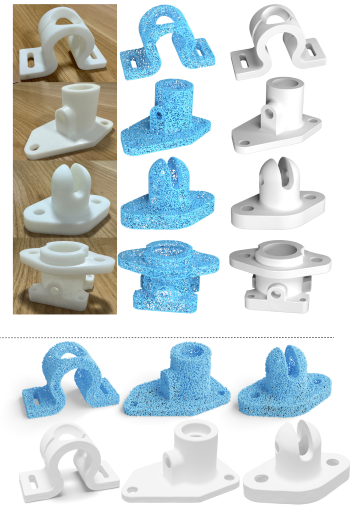}
    \put( 5,35){\textbf{Real Objects}}
    \put(25,35){\textbf{Point Clouds}}
    \put(48,35){\textbf{Our Results}}
    \put(0, 13){\rotatebox{90}{\textbf{Input w/ noise}}}
    \put(0, 4){\rotatebox{90}{\textbf{Ours}}}
    \end{overpic}
    \vspace{-2mm}
    \caption{
    The reconstruction results of our NeurCADRecon on real scanned point clouds.
    Upper: Reconstruction results for point clouds scanned by the SHINING 3D Einscan SE scanner.
    Lower: By adding 0.5\% extra Gaussian noise to the input point clouds, our algorithm demonstrates desirable noise resistance.}
    \label{fig:real_scan}
    \vspace{-3em}
\end{figure}
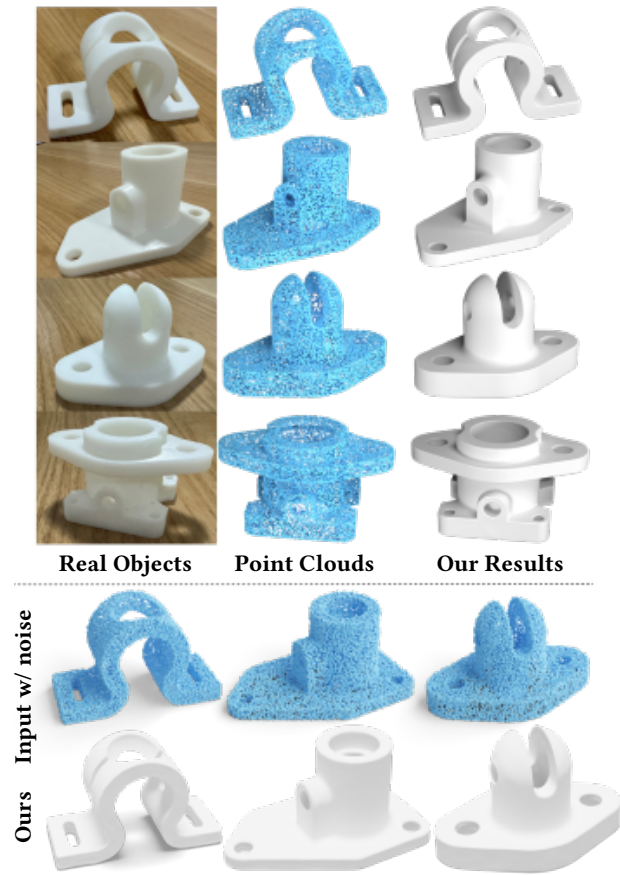

\paragraph{Results on Real Scans} 
In the field of reverse engineering, reconstructing real CAD scans is a pivotal task. To this end, we also evaluate our method's performance on real scanned CAD models, referenced from~\citet{Rui2022RFEPS}. The point clouds generated by the scanner, specifically captured using a SHINING 3D Einscan SE scanner with an accuracy of 0.1mm, present challenges such as noise and nonuniform density.
Benefiting from our Gaussian curvature constraints and dynamic sampling strategy, our method can handle these challenges, preserving feature lines while reconstructing high-fidelity CAD models.
For our analysis, the scanned point clouds contain over 100K points, with some including 400K points, so we randomly selected 20K points to serve as input to our network for the fitting process.
As depicted in the upper part of Fig.~\ref{fig:real_scan}, our method demonstrates a remarkable ability to recover fine details and accurately render concave regions of shapes, showcasing its effectiveness in dealing with real-world CAD model reconstructions.
In the bottom of Fig.~\ref{fig:real_scan}, we present the reconstruction results on the input point clouds after introducing 0.5\% extra Gaussian noise. Notably, our method demonstrates robustness to noise, consistently yielding faithful reconstructions of CAD models.

\subsection{Applications}

\begin{figure}[t]
    \centering
    \begin{overpic}[width=0.95\linewidth]{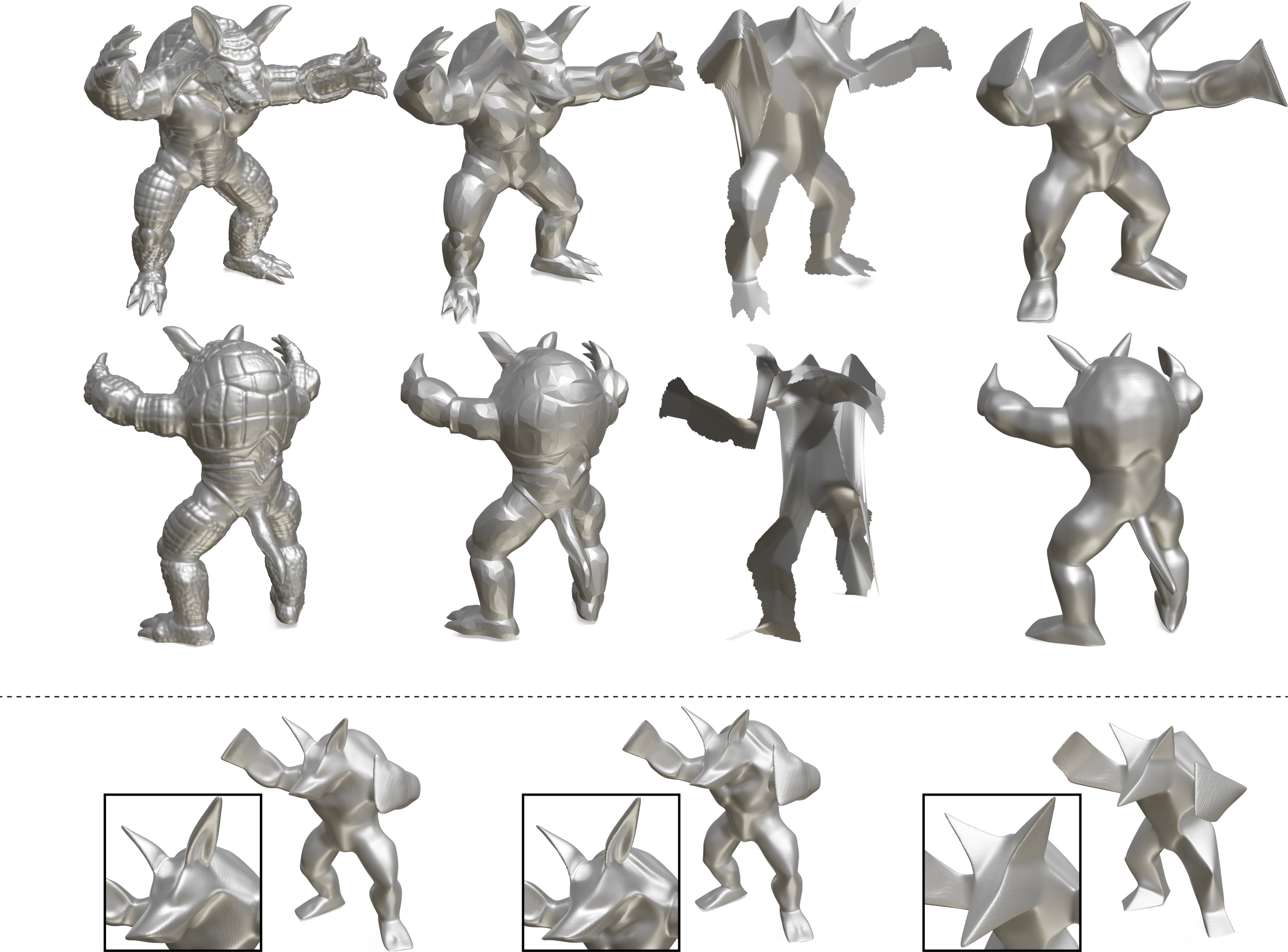}
    \put(0, 38){\rotatebox{90}{\textbf{Comparison}}}
    \put(0, 3){\rotatebox{90}{\textbf{Ours}}}
    \put(13, 21){\textbf{Input}}
    \put(31, 21){\textbf{Stein et al.}}
    \put(54, 21){\textbf{Sell\'{a}n et al.}}
    \put(82, 21){\textbf{Ours}}
    \put(18, -5){\textbf{(a)}}
    \put(51, -5){\textbf{(b)}}
    \put(82, -5){\textbf{(c)}}
    \end{overpic}
    \vspace{2mm}
    \caption{
    	The upper part of this figure presents the results achieved by our proposed method and those reported in~\citet{Develop2018} and~\citet{ Develop2020} for the Armadillo model. The lower part of the figure showcases the results generated by our method under different parameter settings: (a) $\lambda_\text{Gauss}=3$, \#iter 1K, (b) $\lambda_\text{Gauss}=3$, \#iter 7K, (c) $\lambda_\text{Gauss}=100$, \#iter 7K.
    }
    \label{fig:developability}
    \vspace{-2em}
\end{figure}

\begin{figure}[t]
    \centering
    \begin{overpic}[width=0.95\linewidth]{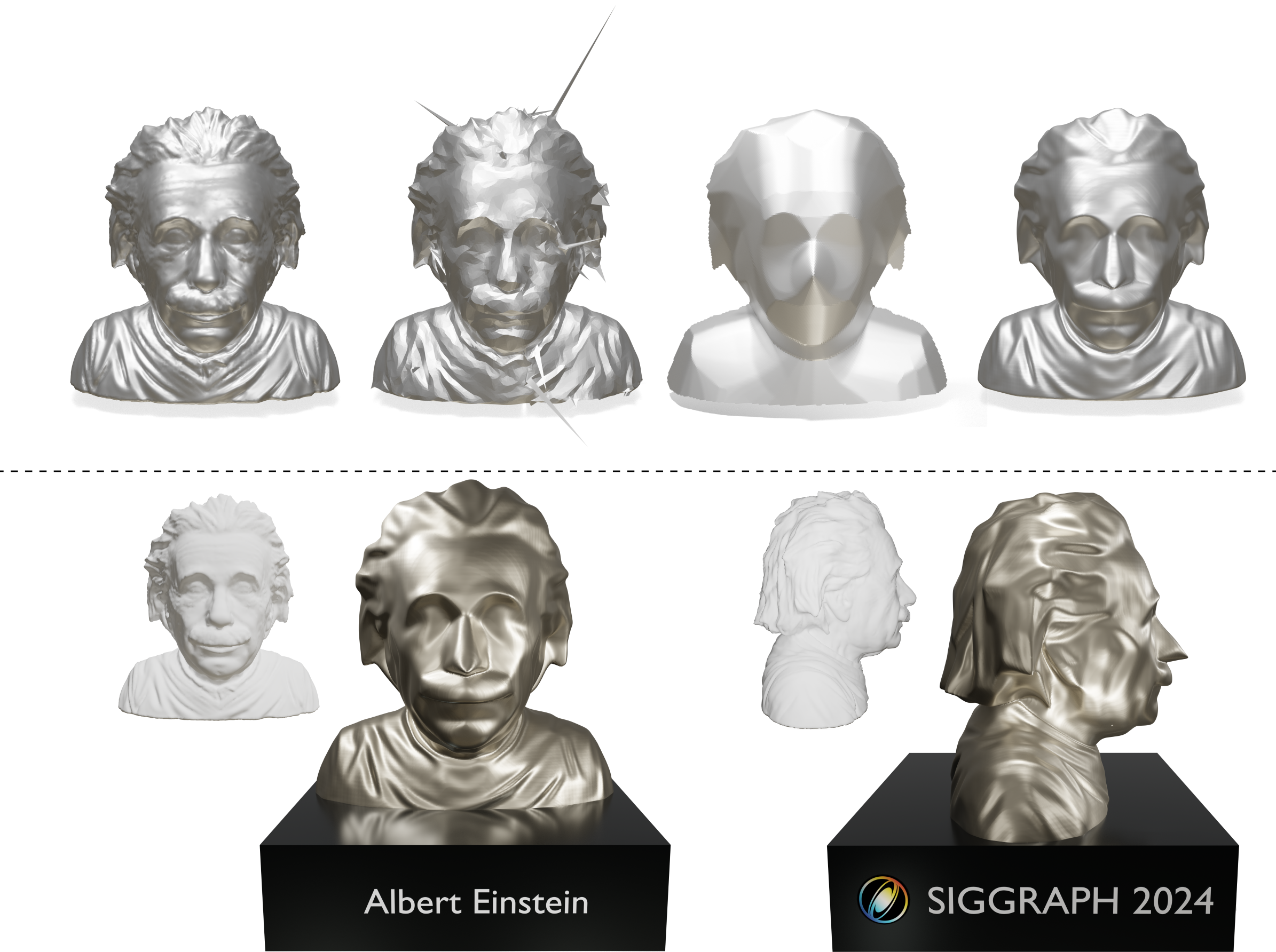}
    \put(0, 53){\textbf{(a)}}
    \put(0, 15){\textbf{(b)}}
    \put(12, 39){\textbf{Input}}
    \put(31, 39){\textbf{Stein et al.}}
    \put(54, 39){\textbf{Sell\'{a}n et al.}}
    \put(83, 39){\textbf{Ours}}
    \end{overpic}
    \vspace{-1mm}
    \caption{
    (a) The results achieved by~\citet{Develop2018}, ~\citet{Develop2020}, and our proposed method for the Einstein model.
    (b) Our approach is well-suited for creating visually appealing artistic characters, thanks to its ability to enforce developability effectively.
    }
    \label{fig:einstein}
    \vspace{-2em}
\end{figure}
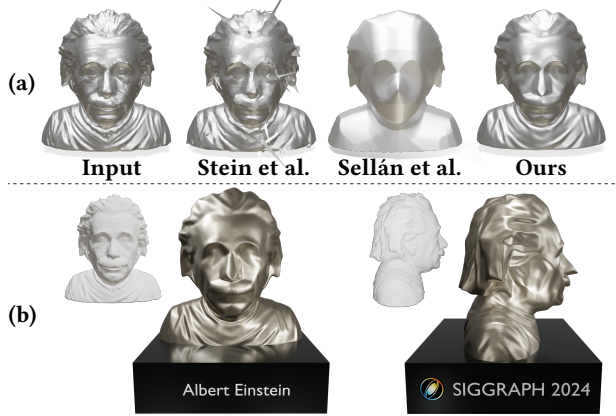

\paragraph{Developability Enforcement}
D-Charts~\cite{DCharts2005} paved the way by introducing a straightforward and robust algorithm for mesh segmentation into developable charts. In contrast, our approach focuses on regulating the Gaussian curvature to guide surface deformation towards approximate developability.
To ensure the sustained impact of the Gaussian curvature term during optimization, we maintain the annealing factor at 1. In the upper part of Fig.~\ref{fig:developability} and Fig.~\ref{fig:einstein}(a), we showcase qualitative outcomes from our method alongside those from the SOTA methods~\cite{Develop2018, Develop2020}. 
In contrast to~\citet{Develop2020}, which often results in an open surface lacking the back side due to its view-dependent nature, our method produces a watertight manifold polygonal surface.

\citet{Develop2018} operates directly on a triangle mesh. Despite its competitive results, the algorithm heavily relies on the mesh tessellation/resolution, and thus cannot consistently produce reliable outputs, as depicted in the upper part of Fig.~\ref{fig:developability} and~\ref{fig:einstein}(a). Our algorithm, independent of mesh tessellation and resolution, can better control the property of developability. Furthermore, for the Armadillo model shown in Fig.~\ref{fig:developability}, our algorithm runs 200 times faster than~\citet{Develop2018}.

Additionally, as illustrated in the lower part of Fig.~\ref{fig:developability}, it's clear that employing more iterations and a higher weight for $\lambda_{\text{Gauss}}$ facilitates the creation of more developable surfaces. Owing to this distinctive feature, our method is particularly adept at crafting artistic characters by effectively enforcing developability, as exemplified in Fig.~\ref{fig:einstein}(b).

\begin{figure}[t]
    \centering
    \begin{overpic}[width=0.95\linewidth]{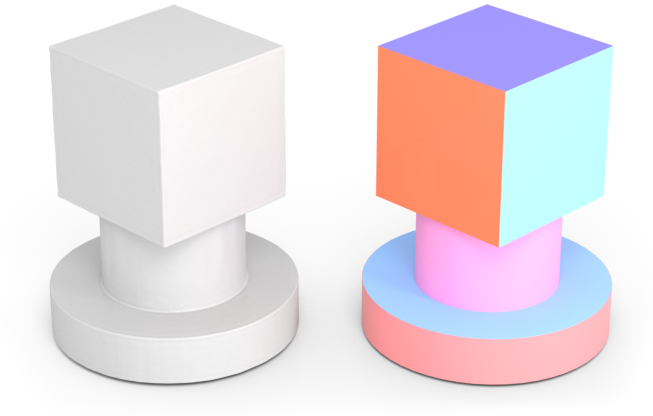}
    \put(15, 0){\textbf{Reconstruction}}
    \put(58, 0){\textbf{Decomposed Patches}}
    \end{overpic}
    \vspace{-2mm}
    \caption{
    The reconstructed surface from our method can be decomposed into distinct smooth surface patches along the sharp feature line, significantly simplifying the challenge of recovering the parametric CAD design.
    }
    \label{fig:patch}
\end{figure}

\paragraph{Design Recovery}
Our algorithm's capability to accurately capture underlying sharp feature points and lines enables straightforward extraction of a feature-aligned triangle mesh from the SDF, as per~\citet{DMC}. Subsequently, this mesh can be decomposed into smooth surface patches, as shown in Fig.~\ref{fig:patch}. Users can then fit a detailed implicit representation to each surface patch, following the methodology outlined by~\citet{halfspaces}. This functionality proves highly beneficial in secondary design, facilitating a variety of post-edit tasks, such as resizing the CAD model.

%% file: sections/05-ablation.tex
\section{Ablation Studies}
\label{sec:ablation}

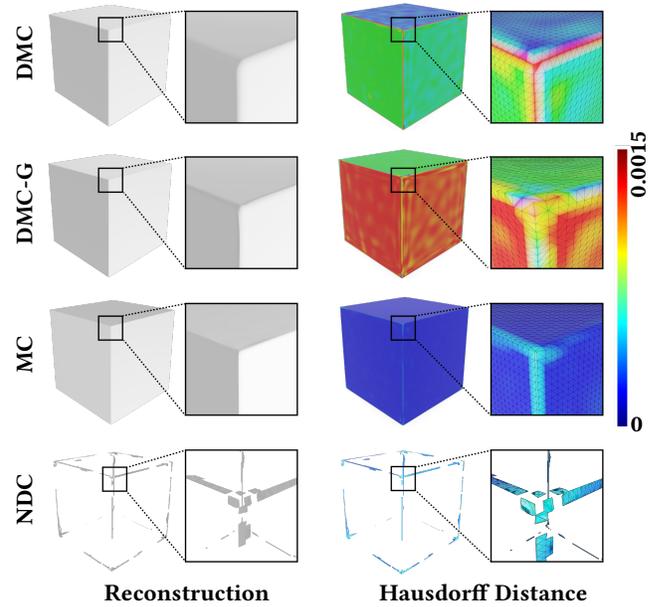
\begin{figure}[t]
    \centering
    \begin{overpic}[width=0.91\linewidth]{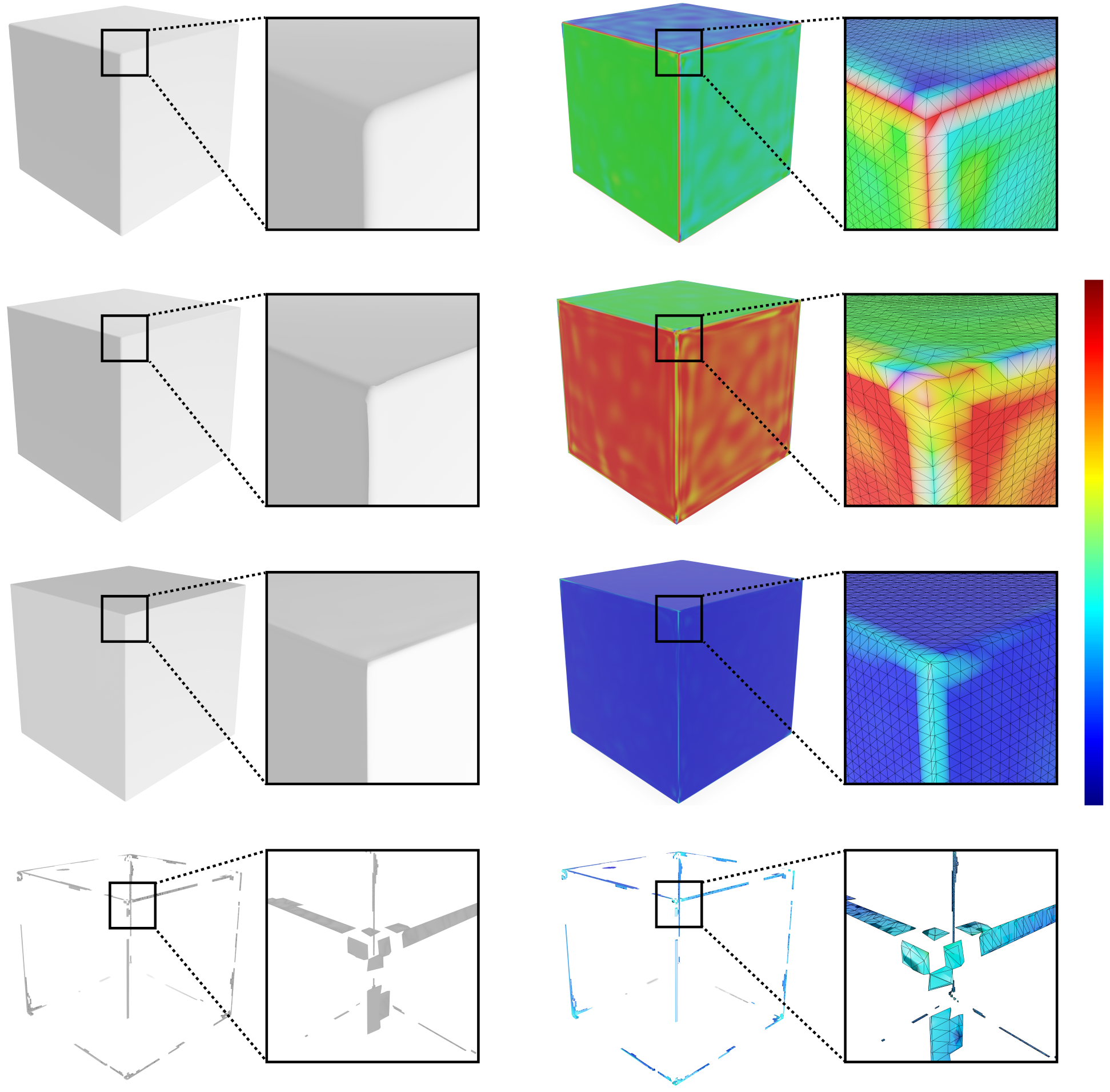}
    \put( -5, 85){\rotatebox{90}{\textbf{DMC}}}
    \put( -5, 57){\rotatebox{90}{\textbf{DMC-G}}}
    \put( -5, 35){\rotatebox{90}{\textbf{MC}}}
    \put( -5, 9){\rotatebox{90}{\textbf{NDC}}}
    \put( 10, -4){\textbf{Reconstruction}}
    \put(57, -4){\textbf{Hausdorff Distance}}
    \put(100, 25){\textbf{0}}
    \put(100, 65){\rotatebox{90}{\textbf{0.0015}}}
    \end{overpic}
    \vspace{1mm}
    \caption{
    The zero isosurface extraction results of Neural Dual
    Contouring (NDC), Marching Cubes (MC), Dual Marching Cubes with gradient (DMC-G), and Dual Marching Cubes without gradient (DMC) for our predicted SDF.
    }
    \label{fig:mcVSdmc}
    \vspace{-2em}
\end{figure}

\subsection{Isosurface Extraction}
\label{sec:mcVSdmc}
In this paper, we utilize the Marching Cubes (MC) algorithm~\cite{lewiner2003efficient} for isosurface extraction. However, the marching cubes algorithm lacks flexibility, as its vertices consistently align to a fixed lattice. Consequently, the generated meshes may fail to align with non-axis-aligned sharp features. On the other hand, Dual Contouring (DC)~\cite{DC_Ju} is popular for its ability to capture sharp features, but it cannot be differentiated. As an improved method, Dual Marching Cubes (DMC)~\cite{DMC} leverages the benefits from both Marching Cubes and Dual Contouring.
Recently, Neural Marching Cubes~(NMC)~\cite{NMC2021} and Neural Dual Contouring~(NDC)~\cite{NDMC2022} have introduced a data-driven approach to positioning the extracted mesh as a function of the input field.

In this experiment, we apply the SDF fields predicted by our method with the double-trough function to both the MC and DMC algorithms. Notably, DMC is adept at handling scenarios with and without gradient fields. We present the reconstruction results for both cases, where the gradient is calculated using the autograd method in Pytorch. In Fig.~\ref{fig:mcVSdmc}, we concurrently showcase the four reconstruction outcomes alongside the computed Hausdorff distance between the reconstructed mesh and the ground truth.

Visual inspection in the left segment of Fig.~\ref{fig:mcVSdmc} reveals that the polygonal surfaces reconstructed by MC and gradient-based DMC exhibit sharper features at corner positions compared to DMC without gradients. Further scrutiny through the zoomed-in visualization on the right side of Fig.~\ref{fig:mcVSdmc} indicates that MC and DMC without gradients struggle to fully capture sharp features at edges and corners. While DMC with gradients can express sharp edges, it still falls short in accurately representing corner shapes. Moreover, a comparison of Hausdorff distances highlights that the reconstruction result of MC is closer to the ground truth. For DMC, both with and without gradients for polygonal surface reconstruction, varying degrees of depressions are observed. In addition, NDC encounters difficulties in accurately extracting the zero isosurface due to divergence issues during optimization. Therefore, we employ the MC algorithm for the extraction of zero isosurfaces in all experiments conducted in this paper.

\begin{table}[!tp]
\vspace{1.5mm}
\centering
\caption{
Comparison of different smooth energy forms: Dirichlet Energy ($E_D$), Hessian Energy ($E_{H_2}$), and $L_1$-based Hessian Energy ($E_{H_1}$). The \under{best} scores are highlighted in bold with underlining.
}
\label{tab:enery}
\begin{tabular}{l|cccccc} 
\toprule
          & \multicolumn{2}{c}{NC~$\uparrow$}  & \multicolumn{2}{c}{CD~$\downarrow$} & \multicolumn{2}{c}{F1~$\uparrow$}  \\
          & mean           & std.          & mean          & std.                     & mean           & std.                                 \\ 
\midrule
$E_{D}$ & 95.24          & 2.92          & 4.11          & 1.96                     & 80.19          & 17.82                                \\
$E_{H_2}$   & 97.54          & 2.01          & 3.11          & 1.61                     & 88.31          & 14.31                                \\
$E_{H_1}$  & 97.22          & 2.25          & 3.06          & 1.75                     & 88.59          & 14.01  \\
\hline
\textbf{Ours}      & \under{98.21} & \under{1.98} & \under{3.01} & \under{1.51}            & \under{91.52} & \under{13.08}                       \\
\bottomrule
\end{tabular}
\end{table}

\begin{figure}[tp]
    \vspace{3mm}
    \centering
    \begin{overpic}[width=0.91\linewidth]{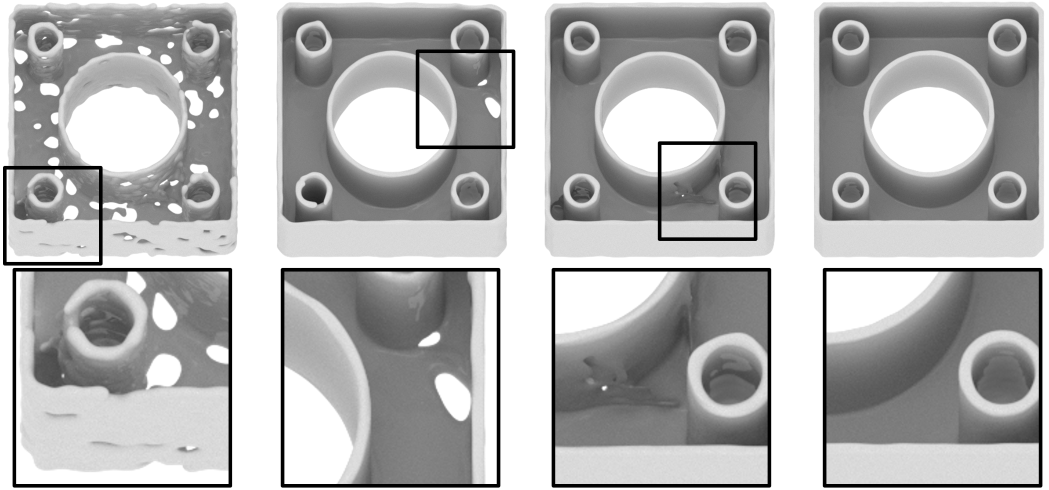}
    \put( 10, -4){\textbf{$E_D$}}
    \put( 35, -4){\textbf{$E_{H_2}$}}
    \put(60, -4){\textbf{$E_{H_1}$}}
    \put(85, -4){\textbf{Ours}}
    \end{overpic}
    \vspace{1mm}
    \caption{
    In a visual comparison of different smooth energy forms, our method demonstrates superiority over other approaches in faithfully recovering the CAD surface without undesirable genus and redundant faces.
    }
    \label{fig:smooth}
\end{figure}
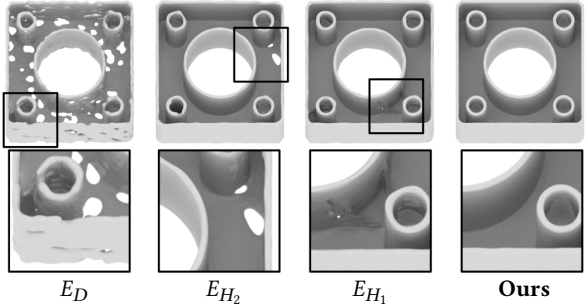

\subsection{Comparison with Other Smoothness Forms}
\label{sec:smooth_fun}
From an algebraic perspective, our Gaussian curvature term can be interpreted as a second-order smooth energy term. Consequently, we compare it with three commonly used smooth energy terms.
They are respectively Dirichlet energy
\begin{equation}
\label{eq:dirichlet_energy}
    E_D(\boldsymbol{x}) = \frac{1}{2}\int_{\mathcal{P}\cup\mathcal{Q}}{\| \nabla f(\boldsymbol{x};\Theta) \|_2^2}dx,
\end{equation}
Hessian energy
\begin{equation}
\label{eq:hessian_energy}
    E_{H_2}(\boldsymbol{x}) = \int_{\mathcal{P}\cup\mathcal{Q}}{\| H_f(\boldsymbol{x}) \|_2^2}dx,
\end{equation}
$L_1$-based Hessian energy~\cite{zhang2022critical}
\begin{equation}
\label{eq:hessian_energy_L1}
    E_{H_1}(\boldsymbol{x}) = \int_{\mathcal{P}\cup\mathcal{Q}}{\| H_f(\boldsymbol{x}) \|_1}dx,
\end{equation}
where $\|\cdot\|_i$ represents the $L_i$ norm.
Note that the Laplacian energy based on the Hessian matrix used by DiGS~\cite{DiGS} is indeed a commonly employed smoothing function, which is also based on SIREN~\cite{SIREN} like ours. Since we have conducted a detailed comparison with DiGS~\cite{DiGS} in the experimental chapter, we will not delve into the details here.

We randomly selected 100 CAD models from all four datasets for experimental comparison, where each CAD model randomly sampled 10K points as input. Tab.~\ref{tab:enery} demonstrates that, for the reconstruction of CAD models, our energy function surpasses the other three energy functions in terms of reconstruction accuracy, normal consistency, and Chamfer distance. Moreover, Fig.~\ref{fig:smooth} showcases the superior reconstruction results of our method, highlighting the undesirable genus in the polygonal surface reconstructed by Dirichlet energy and Hessian energy. Additionally, there are redundant faces in the reconstruction result of $L_1$-based Hessian energy.

\begin{table}[!tp]
\vspace{1.5mm}
\centering
\caption{Ablation studies on the Gaussian curvature constraint term~(GCCT).
In each column, the \under{best} scores are highlighted in bold with underlining.
}
\label{tab:no_gauss_term}
 \resizebox{\linewidth}{!}{%
\begin{tabular}{l|cccccc} 
\toprule
\multirow{2}{*}{}     & \multicolumn{2}{c}{NC~$\uparrow$}  & \multicolumn{2}{c}{CD~$\downarrow$} & \multicolumn{2}{c}{F1~$\uparrow$}  \\
  &   mean           & std.          & mean          & std.                     & mean           & std.                  \\ 
\midrule
w/o GCCT  
& 68.29   & 3.57 & 9.34 & 6.84 & 75.69 & 18.62 \\
w/ GCCT (\textbf{Ours})  
& \under{98.65}   & \under{1.96} & \under{2.19} & \under{1.21} & \under{91.58} & \under{11.36} \\
\bottomrule
\end{tabular}
}
\end{table}

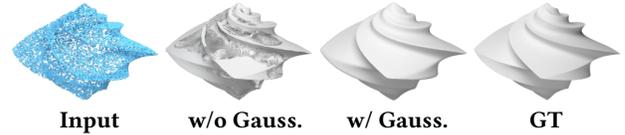
\begin{figure}[t]
    \vspace{3mm}
    \centering
    \begin{overpic}[width=0.95\linewidth]{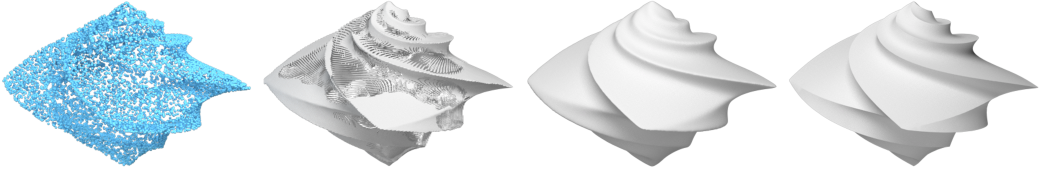}
    \put( 8, -4){\textbf{Input}}
    \put( 29, -4){\textbf{w/o Gauss.}}
    \put(55, -4){\textbf{w/ Gauss.}}
    \put(85, -4){\textbf{GT}}
    \end{overpic}
    \vspace{1mm}
    \caption{
    Reconstruction results are presented both without~(w/o Gauss.) and with~(w/ Gauss.) the Gaussian curvature term.
    }
    \label{fig:flower}
\end{figure}

\subsection{W/WO Gaussian Curvature Term}
\label{sec:no_gauss_term}
To further emphasize the efficacy of our Gaussian curvature loss term in CAD model reconstruction, we conducted a comparative analysis by disabling the Gaussian term.

We randomly selected 100 models for testing and comparison and then set the Gaussian curvature weight term~$\lambda_{\text{Gauss}}$ to 0, leaving the other settings unchanged.
The compelling performance observed in the quantitative results presented in Tab.~\ref{tab:no_gauss_term} underscores the superiority of our method in reconstructing CAD models, providing further evidence of its effectiveness.
The qualitative results in Fig.~\ref{fig:flower} indicate that when reconstructing a complex CAD model without utilizing the Gaussian curvature term, the reconstruction results will have an undesirable genus.

\begin{table}[tp]
\vspace{1.5mm}
\centering
\caption{A comparative analysis of timing costs and the number of parameters~(\#Para.) per iteration for IGR~\cite{IGR}, SIREN~\cite{SIREN}, DiGS~\cite{DiGS}, and NSH~\cite{HessianZX}, all executed without the supervision of normals. The timing statistics are meticulously reported in milliseconds (ms).
}
\label{tab:runtime}
\begin{tabular}{l|ccccc} 
\toprule
              & IGR & SIREN & DiGS & NSH & Ours  \\
\midrule
\#Para. & 1.86M     &  264.4K     & 264.4K  & 264.4K & 264.4K      \\
time~[ms]     & 50.73      &  11.52   &  36.28    & 40.10 & 39.58     \\
\bottomrule
\end{tabular}
\end{table}

\subsection{Runtime Performance}
\label{sec:runtime}
We compare IGR~\cite{IGR}, SIREN~\cite{SIREN}, DiGS~\cite{DiGS}, NSH~\cite{HessianZX}, and our method in terms of runtime performance.
These statistical insights are derived from our experimental platform, as detailed in Sec.~\ref{sec:exp_setting}.
For all approaches, we set the batch size to 10K and the number of hidden layers to 4, 
with each layer containing 256 units.
Tab.~\ref{tab:runtime} reports the time costs (in milliseconds) for a single iteration.
DiGS~\cite{DiGS}, NSH~\cite{HessianZX}, and our approach are all based on SIREN~\cite{SIREN}, utilizing identical parameter quantities.
In comparison to SIREN~\cite{SIREN}, these three methods demand the estimation of second-order geometric quantities, leading to uniformly higher time consumption than SIREN.
IGR runs inefficiently due to a different implementation.

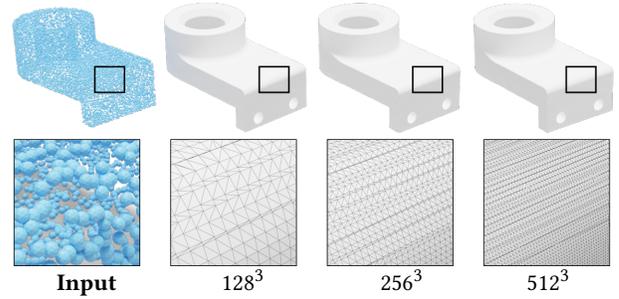
\begin{figure}[t]
    \vspace{3mm}
    \centering
    \begin{overpic}[width=0.95\linewidth]{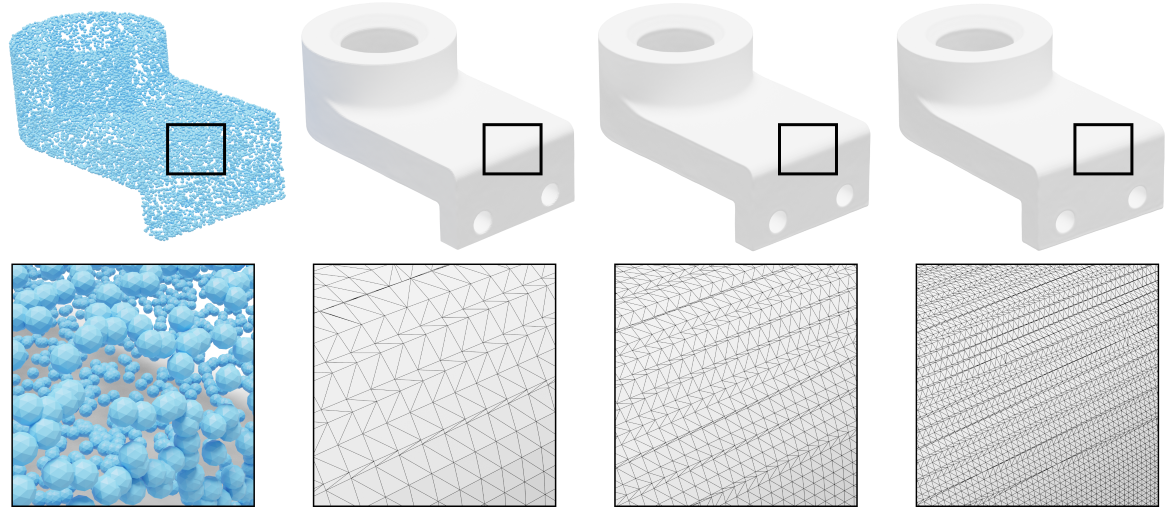}
    \put( 8, -4){\textbf{Input}}
    \put(35, -4){\textbf{$128^3$}}
    \put(61, -4){\textbf{$256^3$}}
    \put(85, -4){\textbf{$512^3$}}
    \end{overpic}
    \vspace{1mm}
    \caption{
     The reconstruction results of our method across different MC~\cite{lewiner2003efficient} resolutions, where the magnification ratio is 12 times.
    }
    \label{fig:diff_mc}
\end{figure}

In Table~\ref{tab:runtime_diff}, we present the runtime performance of our method across various input sizes of point clouds and different Marching Cubes~(MC)~\cite{lewiner2003efficient} resolutions. The data indicate that as the number of input point clouds and the MC~\cite{lewiner2003efficient} resolution increase, the time consumption of our method increases relatively slowly.
In Fig.~\ref{fig:diff_mc}, we present the reconstruction results of our method across different MC~~\cite{lewiner2003efficient} resolutions.

\begin{table}[tp]
\centering
\caption{The runtimes of our method for different input point clouds and resolution of MC~\cite{lewiner2003efficient}, which are reported in milliseconds (ms). We maintained the MC resolution at $256^3$ when presenting the runtime efficiency for various input sizes. Similarly, we utilized 10K input points to discuss the runtime efficiency for different MC resolutions.}
\label{tab:runtime_diff}
\resizebox{0.95\linewidth}{!}{
\begin{tabular}{l|cccc|ccc} 
\toprule
\multirow{2}{*}{} & \multicolumn{4}{c}{Input Size} & \multicolumn{3}{c}{MC Resolution}  \\ 
\cmidrule{2-5} 
\cmidrule{6-8}
   & 10K & 20K & 50K & 80K & $128^3$ & $256^3$ & $512^3$ \\
\cmidrule{1-8}
time~[ms] & 39.58 & 46.18 & 92.35 & 145.13 & 9.15 & 39.58 & 293.12\\
\bottomrule
\end{tabular}
}
\end{table}

%% file: sections/06-conclusion.tex
\section{Limitation}
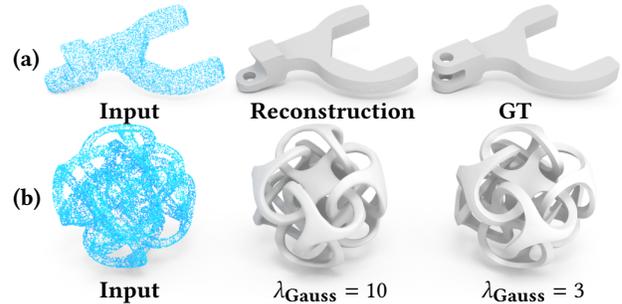
\begin{figure}[t]
    \vspace{3mm}
    \centering
    \begin{overpic}[width=0.9\linewidth]{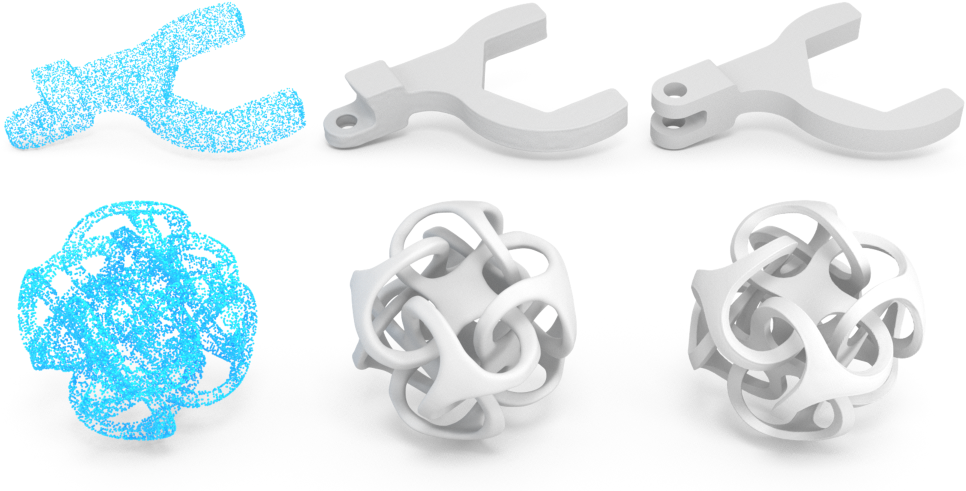}
    \put(-5, 40){\textbf{(a)}}
    \put(-5, 16){\textbf{(b)}}
    \put(10, 31){\textbf{Input}}
    \put(36, 31){\textbf{Reconstruction}}
    \put(79, 31){\textbf{GT}}
    \put(10, 0){\textbf{Input}}
    \put(40, 0){\textbf{$\lambda_\text{Gauss}=10$}}
    \put(76, 0){\textbf{$\lambda_\text{Gauss}=3$}}
    \end{overpic}
    \vspace{-2mm}
    \caption{
    (a) NeurCADRecon may encounter challenges in cases with a substantial number of missing points. (b) Reconstruction outcomes of a complex model using our method under varying weight conditions.
    }
    \label{fig:limitation}
\end{figure}

Our current approach has at least two drawbacks, as seen in Fig.~\ref{fig:limitation}. Firstly, when the missing parts of the input point cloud are too large, our approach cannot guarantee the recovery of the actual shape. Secondly, although we use the same parameter to weigh the influence of Gaussian curvature for all experiments in this paper, we notice that there may be some special cases where the parameter needs fine-tuning since the extent to enforce zero Gaussian curvature is related to the inherent geometric complexity.

\section{Conclusion}
In this paper, we introduce NeurCADRecon, a self-supervised approach for neural signed distance function estimation. Our method is designed to work with unoriented point cloud inputs representing CAD models. We leverage prior knowledge that CAD model surfaces are typically composed of piecewise smooth patches, each of which is approximately developable. To achieve this, we minimize the overall absolute Gaussian curvature. To accommodate scenarios where the Gaussian curvature has non-zero values at tip points, we introduce a double-trough curve. Additionally, we propose a dynamic sampling strategy to handle data imperfections. Extensive tests on public datasets demonstrate that NeurCADRecon outperforms existing state-of-the-art methods in reconstructing high-fidelity CAD shapes.